\documentclass[10pt,twocolumn,letterpaper]{article}

\usepackage{iccv}      

\usepackage{bm}
\usepackage{algorithm}
\usepackage{algorithmic}
\usepackage{multirow}
\usepackage{graphicx}
\usepackage{times}
\usepackage{epsfig}
\usepackage{graphicx}
\usepackage{amsmath}
\usepackage{amssymb}
\usepackage{caption}
\usepackage{subcaption}
\usepackage{interval}
\usepackage{booktabs}
\usepackage{multirow}
\usepackage{textcmds}

\usepackage{colortbl}
\usepackage{tabulary}
\usepackage{etoolbox}
\usepackage[dvipsnames]{xcolor}

\definecolor{tabfirst}{rgb}{1, 0.7, 0.7} 
\definecolor{tabsecond}{rgb}{1, 0.85, 0.7}
\definecolor{tabthird}{rgb}{1, 1, 0.7} 
\newcommand{\best}[2]{%
  \ifnum#1=1
    \cellcolor{red!50}#2%
  \else\ifnum#1=2
    \cellcolor{orange!50}#2%
  \else\ifnum#1=3
    \cellcolor{yellow!50}#2%
  \else
    #2%
  \fi\fi\fi
}

\definecolor{iccvblue}{rgb}{0.21,0.49,0.74}
\usepackage[pagebackref,breaklinks,colorlinks,allcolors=iccvblue]{hyperref}

\def\paperID{12526} 
\def\confName{ICCV}
\def\confYear{2025}

\title{GeoDiff: Geometry-Guided Diffusion for Metric Depth Estimation}

\author{
Tuan Pham\textsuperscript{*} \quad
Thanh-Tung Le\textsuperscript{*} \quad
Xiaohui Xie\textsuperscript{\dag} \quad
Stephan Mandt\textsuperscript{\dag} \\
University of California, Irvine \quad \textsuperscript{*} Equal contribution \quad \textsuperscript{\dag} Equal advising \\
{\tt\small \{tuan.pham, tung.le, xhx, mandt\}@uci.edu}
}

\begin{document}

\twocolumn[{
\renewcommand\twocolumn[1][]{#1}
\maketitle
\centering
\captionsetup{type=figure}\includegraphics[width=\linewidth]{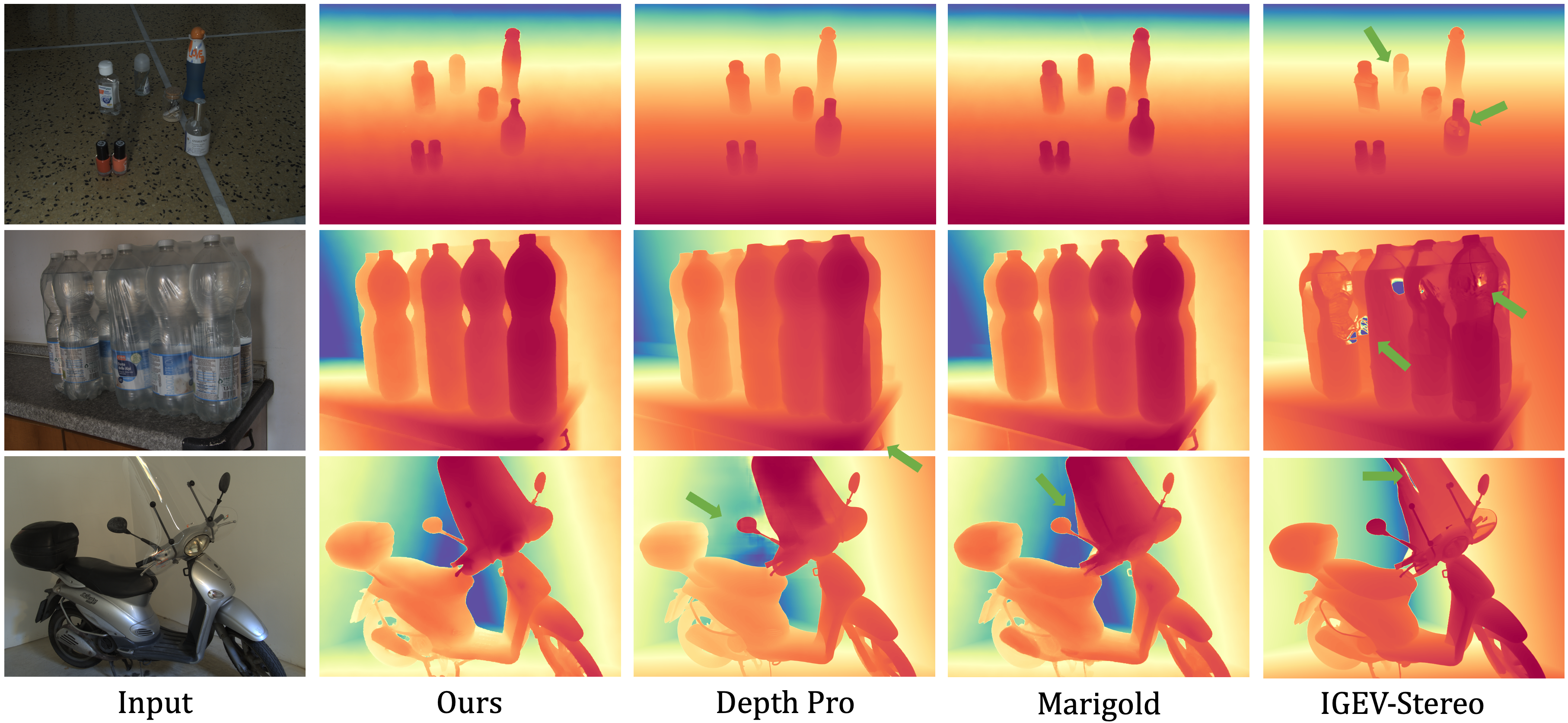}
\captionof{figure}{
\textbf{Results on the challenging Booster~\cite{zamaramirez2024booster} dataset.} Comparison of depth estimation performance on transparent and reflective objects. From left to right: input image, followed by depth maps from GeoDiff (ours), DepthPro~\cite{bochkovskii2024depth}, Marigold~\cite{ke2024repurposing}, and IGEV-Stereo~\cite{xu2023iterative}. Our method generates sharp, accurate metric depth maps in a zero-shot setting, leveraging stereo pairs for improved depth recovery.
}
\label{fig:demo}
}]

\maketitle

\begin{abstract}
We introduce a novel framework for metric depth estimation that enhances pretrained diffusion-based monocular depth estimation (DB-MDE) models with stereo vision guidance. While existing DB-MDE methods excel at predicting relative depth, estimating absolute metric depth remains challenging due to scale ambiguities in single-image scenarios. To address this, we reframe depth estimation as an inverse problem, leveraging pretrained latent diffusion models (LDMs) conditioned on RGB images, combined with stereo-based geometric constraints, to learn scale and shift for accurate depth recovery. Our training-free solution seamlessly integrates into existing DB-MDE frameworks and generalizes across indoor, outdoor, and complex environments. Extensive experiments demonstrate that our approach matches or surpasses state-of-the-art methods, particularly in challenging scenarios involving translucent and specular surfaces, all without requiring retraining.
\end{abstract}    

\section{Introduction}
\label{sec:intro}
Depth estimation is an essential task and play a fundamental role in wide applications, such as 3D reconstruction~\cite{mildenhall2021nerf, kerbl20233d}, autonomous driving~\cite{wang2019pseudo}, and AI-generated content~\cite{zhang2023adding, liew2023magicedit}. Recently, monocular depth estimation (MDE)~\cite{birkl2023midas, eigen2014depth, fu2018deep, saxena2023monocular} and stereo depth estimation (StDE) have emerged as the leading methods for depth estimation. MDE approaches generally predict relative depth, which is invariant to scale and shift, whereas StDE methods focus on predicting disparity between two input images, which can be converted to metric depth (in meters) using known camera baseline and focal length. While recent methods attempt direct metric depth prediction from monocular images through large foundation models~\cite{yin2023metric3d, yang2024depth2, yang2024depth1}, these approaches demand extensive synthetic and real data and are computationally expensive to train. 

Recently, diffusion models have demonstrated potential as robust priors for zero-shot dense prediction tasks, including depth estimation~\cite{duan2023diffusiondepth, ke2024repurposing, ji2023ddp, saxena2024surprising, saxena2023monocular, fu2024geowizard, patni2024ecodepth}. Marigold~\cite{ke2024repurposing} is among one of the pioneer methods that propose to repurpose diffusion-based image generators for MDE. The main idea is to finetune pretrained latent diffusion models (LDMs)~\cite{rombach2022high}, which have been trained on extensive text-to-image datasets, to generate depth maps from noise by conditioning on RGB images. The simple diffusion-based MDE (DB-MDE) paradigm works surprisingly well, delivering strong performance across a diverse range of natural images. Subsequent studies~\cite{fu2024geowizard, he2024lotus, gui2024depthfm, duan2023diffusiondepth} have developed upon this paradigm, establishing diffusion-based MDE as an active research in dense prediction task. However, estimating metric depth from a single image remains an inherently ill-posed and challenging problem, leading most DB methods to concentrate on reconstructing relative depth rather than absolute metric depth. 

\begin{figure}[t!]
\centering
\includegraphics[width=0.47\textwidth]{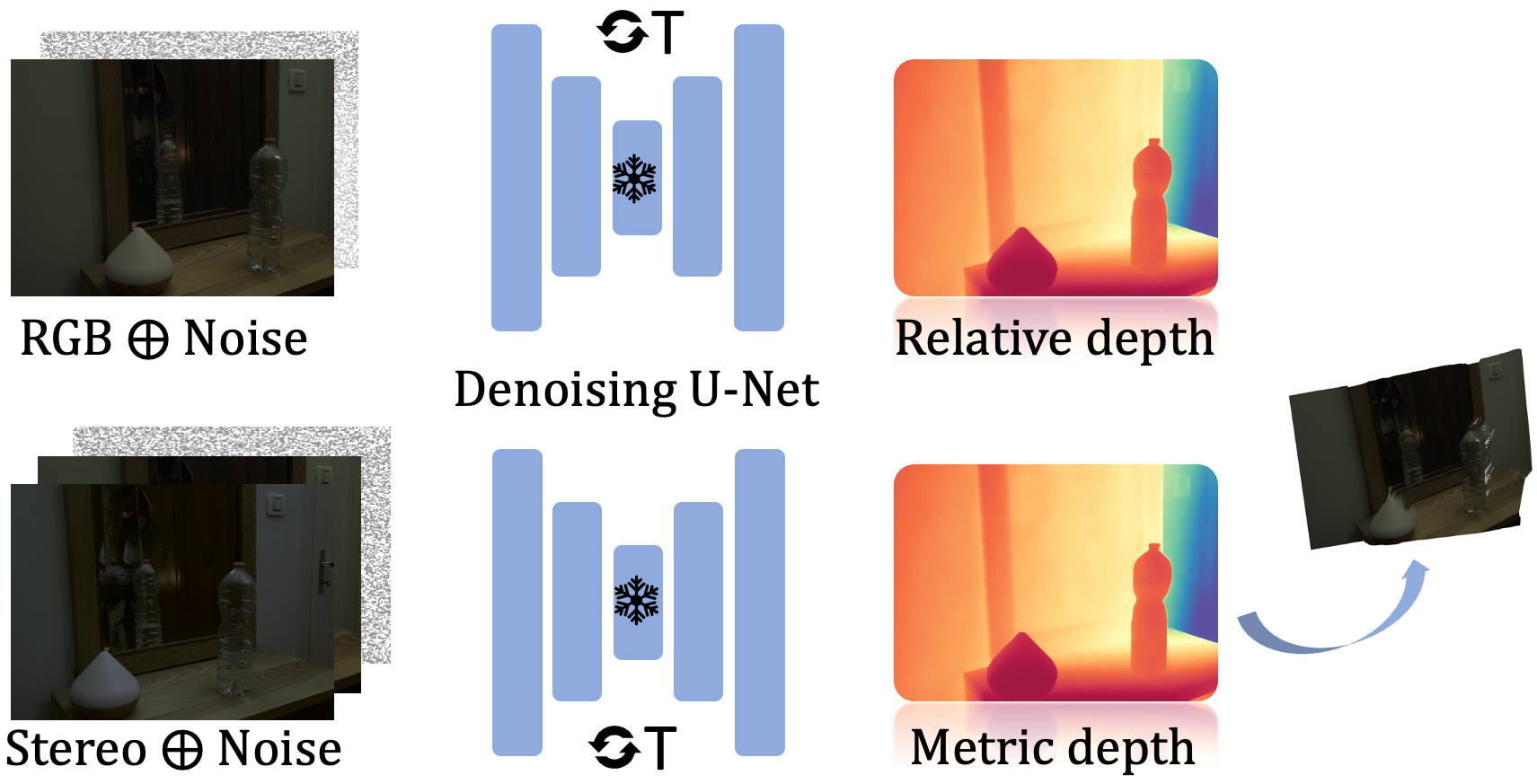}
\caption{\textbf{Top.} Prior methods~\cite{ke2024repurposing, fu2024geowizard} focus on fine-tuning diffusion models to estimate relative depth. \textbf{Bottom.} In contrast, our approach, without retraining, combines a pretrained monocular model with geometric guidance from stereo cues to directly predict metric depth in meters, achieving superior accuracy in even challenging scenes such as transparent and reflective surfaces.}
\label{fig:teaser}
\vspace{-0.8em}
\end{figure}
In this work, we advance this line of research by leveraging pretrained priors from DB-MDE models to achieve metric depth estimation through the incorporation of additional stereo settings. Specifically, we reformulate the depth sampling process as an inverse problem (IP) solved through diffusion models~\cite{daras2024survey, song2023solving, chung2022diffusion, chung2022improving, kawar2022denoising}.  Our approach leverages pretrained LDMs, along with stereo vision-based geometric guidance, to learn the scale and shift for any given scene. Built upon the foundation of diffusion-based MDE (DB-MDE) approaches, such as Marigold~\cite{ke2024repurposing}, our method is scene-agnostic—applicable to objects, indoor, and outdoor scenes—and can be integrated with any DB-MDE framework following a similar schema. Extensive experiments across diverse datasets demonstrate that our approach performs effectively on a wide range of scenes and depth scales without requiring re-training for specific use cases. In summary, we propose:
\begin{itemize}
    \item A novel framework that leverages diffusion-based MDE priors and stereo settings to achieve metric depth estimation.
    \item An IP-based approach for depth estimation that introduces a plug-and-play module, seamlessly integrating with any pretrained diffusion-based depth models that use iterative updates.
    \item Extensive experimental evaluations of our method compared to other competing methods across various datasets, encompassing indoor, outdoor, and challenging scenes with translucent or specular surfaces. 
\end{itemize}
\section{Related works}
\label{sec:related-works}

\subsection{Depth Estimation}
Depth estimation has various applications in 3D vision~\cite{ke2024repurposing, fu2024geowizard, le2023diffeomorphic, le2024integrating, guo2019group, han2024hybrid, sun2023hybrid}. Monocular depth estimation and stereo matching have both seen significant advancements in recent years. Traditional monocular methods focused on in-domain metric depth estimation but faced challenges in generalization, leading to a shift towards zero-shot relative depth estimation using approaches like Stable Diffusion~\cite{rombach2022high} for depth denoising and large-scale datasets such as ~\cite{ke2024repurposing, fu2024geowizard}. However, these data-driven models are limited by the availability of synthetic dataset, which has prompted methods like DepthAnything~\cite{yang2024depth1,yang2024depth2}, DepthPro~\cite{bochkovskii2024depth} to leverage vast amounts of additional real dataset for enhanced robustness. Stereo matching, on the other hand, relies heavily on cost volume filtering techniques, typically employing deep learning models to extract features, build cost volumes, and regress disparities. Models like GCNet~\cite{kendall2017end} and PSMNet~\cite{chang2018pyramid} use 3D CNN architectures to address challenges with occlusions and textureless surfaces, while newer methods, such as GwcNet~\cite{guo2019group} and ACVNet~\cite{xu2022attention}, introduce group-wise correlation and attention mechanisms to improve cost volume expressiveness. RAFT-Stereo~\cite{lipson2021raft} adapts the optical flow network RAFT~\cite{teed2020raft} with multi-level convolutional GRUs, achieving impressive results. Building on this foundation, IGEV-Stereo~\cite{xu2023iterative} proposes iterative geometry encoding volumes, demonstrating improved robustness.  Despite these advancements, the high memory and computational costs of 3D convolutions limit scalability. Our method leverages diffusion priors and geometric guidance, effectively overcoming the limitations of traditional cost volume approaches, particularly for challenging surfaces like transparent or reflective regions. In particular, our approach employs pretrained model from diffusion-based MDE combining with stereo information to estimate metric depth in the wild. To the best of our knowledge, we are the first to employ a training-free approach that combines both monocular and stereo information to solve metric depth estimation effectively.
\subsection{Diffusion Models for Inverse Problem Solving}
Inverse problems (IPs) are ubiquitous and and associated with a wide range of reconstruction problems such as computational image~\cite{beck2009fast, afonso2010augmented}, medical imaging~\cite{suetens2017fundamentals, ravishankar2019image}, and remote sensing~\cite{liu2021automated}. IPs aim to recover an unknown sample $x \in \mathbb{R}^n$, given observed measurements $y \in \mathbb{R}^m$ of the form: $y = \mathcal{A}(x) + e$, where function $\mathcal{A}(\cdot):\mathbb{R}^n \rightarrow \mathbb{R}^m$ is the forward measurement operator and $e \in \mathbb{R}^m$ is additive noise. In the literature, the traditional approach of using hand-crafted priors (e.g. sparsity) is slowly being replaced by rich, learned priors such as diffusion generative models. While recent works~\cite{song2021solving, chung2022diffusion, chung2022improving} propose to solve inverse problem in pixel space, which is computationally expensive, authors~\cite{chung2022improving, song2023solving} recently introduce a method to solve IP in the latent space. In this work, we demonstrate that by using latent DM and classical stereo vision as geometry guidance, we can solve for metric depth estimation problem without re-training MDE. We hope that our approach potentially opens a new direction for tackling depth estimation problems.

\vspace{-0.5em}
\section{Background}
\label{sec:background}

In this section, we provide background regarding diffusion models for monocular depth estimation in Section~\ref{subsec:DB-MDE}. Then, we describe how to employ diffusion models for solving inverse problems in Section~\ref{subsec:DM-Inverse}. Finally, we demonstrate the differentiable warping module in Section~\ref{subsec:stereo-reproj}.
\subsection{Diffusion models for MDE}
\label{subsec:DB-MDE}
MDE is formulated as a conditional denoising diffusion generation task, modeling $p(\bm{x}|\bm{y})$, where $\bm{x} \in \mathbb{R}^{H \times W \times 1}$ denotes depth and $\bm{y} \in \mathbb{R}^{H \times W \times 3}$ represents RGB input. The forward process progressively perturbs data via Gaussian kernels through a variance-preserving SDE~\cite{song2020score}:
\begin{equation} \label{eq:forward}
   d \bm{x} = - \frac{\beta_t}{2} \bm{x} dt + \sqrt{\beta_t} d \bm{w}
\end{equation}
where $\beta_t \in (0,1)$ is the monotonically increasing noise schedule and $\bm{w}$ denotes standard Wiener process. The reverse process learns the corresponding reverse SDE:
\begin{equation} \label{eq:reverse}
   d \bm{x} = \left[ -\frac{\beta_t}{2} \bm{x} - \beta_t \nabla_{\bm{x}_t} \log p(\bm{x}_t|\bm{y}) \right]dt + \sqrt{\beta_t} d\bar{\bm{w}}
\end{equation}
where $\nabla_{\bm{x}_t} \log p(\bm{x}_t)$ is the score function and $d\bar{\bm{w}}$ denotes backward Wiener process. A denoising score matching network~\cite{vincent2011connection} is trained to approximate the score function: 
\begin{equation}
   \hat{\theta} = \text{argmin}_\theta \mathbb{E} [ || \bm{s}_\theta (\bm{x}_t, \bm{y}, t) - \nabla_{\bm{x}_t} \log p(\bm{x}_t | \bm{x_0}, \bm{y})||_2^2]
\end{equation}
The trained score function $\bm{s}_\theta$ is then used to approximate the reverse-time SDE through numerical simulation.

\subsection{Diffusion Models for Solving Inverse Problems}
\label{subsec:DM-Inverse}
In inverse problems, we can recover an unknown signal $\bm{x}$ from measurements $\bm{y}$ related by $\bm{y} = \mathcal{A}(\bm{x}) + \bm{e}$, where $\mathcal{A}(\cdot)$ is the forward measurement operator and $\bm{e} \sim \mathcal{N}(0, \sigma^2 I)$ represents Gaussian noise \citep{chung2022diffusion}. Applying Bayes' theorem to the conditional score:
\begin{equation}
   \nabla_{\bm{x}_t} \log p(\bm{x}_t \mid \bm{y}) = \nabla_{\bm{x}_t} \log p(\bm{x}_t) + \nabla_{\bm{x}_t} \log p(\bm{y} \mid \bm{x}_t)
\end{equation}
\noindent Under mild assumptions \citep{chung2022diffusion}, we approximate:
\begin{equation}
\nabla_{\bm{x}_t} \log p(\bm{y} \mid \bm{x}_t) \approx \nabla_{\bm{x}_t} \log p(\bm{y} \mid \hat{\bm{x}}_0)
\end{equation}
where $\hat{\bm{x}}_0$ is the one-step prediction via Tweedie's formula~\citep{robbins1992empirical}: 
\begin{equation}
\label{eq:tweedie}
\hat{\bm{x}}_0=\frac{1}{\sqrt{\bar{\alpha}_t}}\left(\bm{x}_t+\sqrt{1-\bar{\alpha}_t} \bm{s}_\theta\left(\bm{x}_t, t, \bm{y}_1 \right)\right)
\end{equation}
\noindent With Gaussian noise, we derive:
\begin{equation}
\nabla_{\bm{x}_t} \log p(\bm{y} \mid \hat{\bm{x}}_0) =  - \frac{1}{\sigma^2} \nabla_{\bm{x}_t} \| \bm{y} - \mathcal{A}(\hat{\bm{x}}_0) \|_2^2
\end{equation}
\noindent The final conditional score becomes:
\begin{equation}
\label{eq:cond_score_inv}
   \nabla_{\bm{x}_t} \log p(\bm{x}_t \mid \bm{y}) = \nabla_{\bm{x}_t} \log p(\bm{x}_t) - \lambda \nabla_{\bm{x}_t} \| \bm{y} - \mathcal{A}(\hat{\bm{x}}_0) \|_2^2
\end{equation}
where $\lambda$ controls the strength of additional guidance to the original score function.

\subsection{Differentiable warping}
\label{subsec:stereo-reproj}
Given a stereo pair (or two views with known poses) $\bm{y}_1$, $\bm{y}_2$ and the corresponding depth maps $\bm{x}_1$, $\bm{x}_2$; we define an operation that projects each pixel from the source image $\bm{y}_1$ onto the target image $\bm{y}_2$. Using the intrinsic matrices $K_1, K_2 \in \mathbb{R}^{3\times3}$ of the source and target cameras, respectively, and the relative transformation $T_{1 \rightarrow 2} \in \mathbb{R}^{4 \times 4}$ between the cameras, the forward warping is formulated as:
\begin{equation}
\label{eq:forward-warp}
\bm{c}_2 \thicksim K_2 T_{1\rightarrow 2}  \bm{x}_1(\bm{c}_1)  K_1^{-1}  \bm{c}_1
\end{equation}
where $\bm{c}_1$ and $\bm{c}_2$ denote the homogeneous pixel coordinates in $\bm{y}_1$ and $\bm{y}_2$, respectively, and $\bm{x}_1(\bm{c}_1)$ represents the depth at pixel $\bm{c}_1$ in $\bm{y}_1$.

Based on this coordinates mapping, we can define the forward warping operator $\bm{P}_{\bm{y}_1 \rightarrow \bm{y}_2} \left( \bm{x}_1, \bm{y}_1 \right)$, which projects $\bm{y}_1$ onto $\bm{y}_2$; and the backward warping operator $\bm{P}_{\bm{y}_2 \rightarrow \bm{y}_1}\left(\bm{x}_1, \bm{y}_2\right)$. Notably, both warping operations rely solely on the depth map $\bm{x}_1$ from $\bm{y}_1$. Backward warping, in particular, has been extensively applied for computing reprojection losses in self-supervised depth estimation \citep{godard2017unsupervised, godard2019digging} or online stereo depth adaptation \citep{tonioni2019learning, zhang2019online}. We provide in-depth discussion in the Supplementary~\ref{subsec:supp_diff_warping}.
\section{Methodology}
\label{sec:method}
\begin{figure*}[t]
  \begin{center}

  \includegraphics[width=\textwidth]{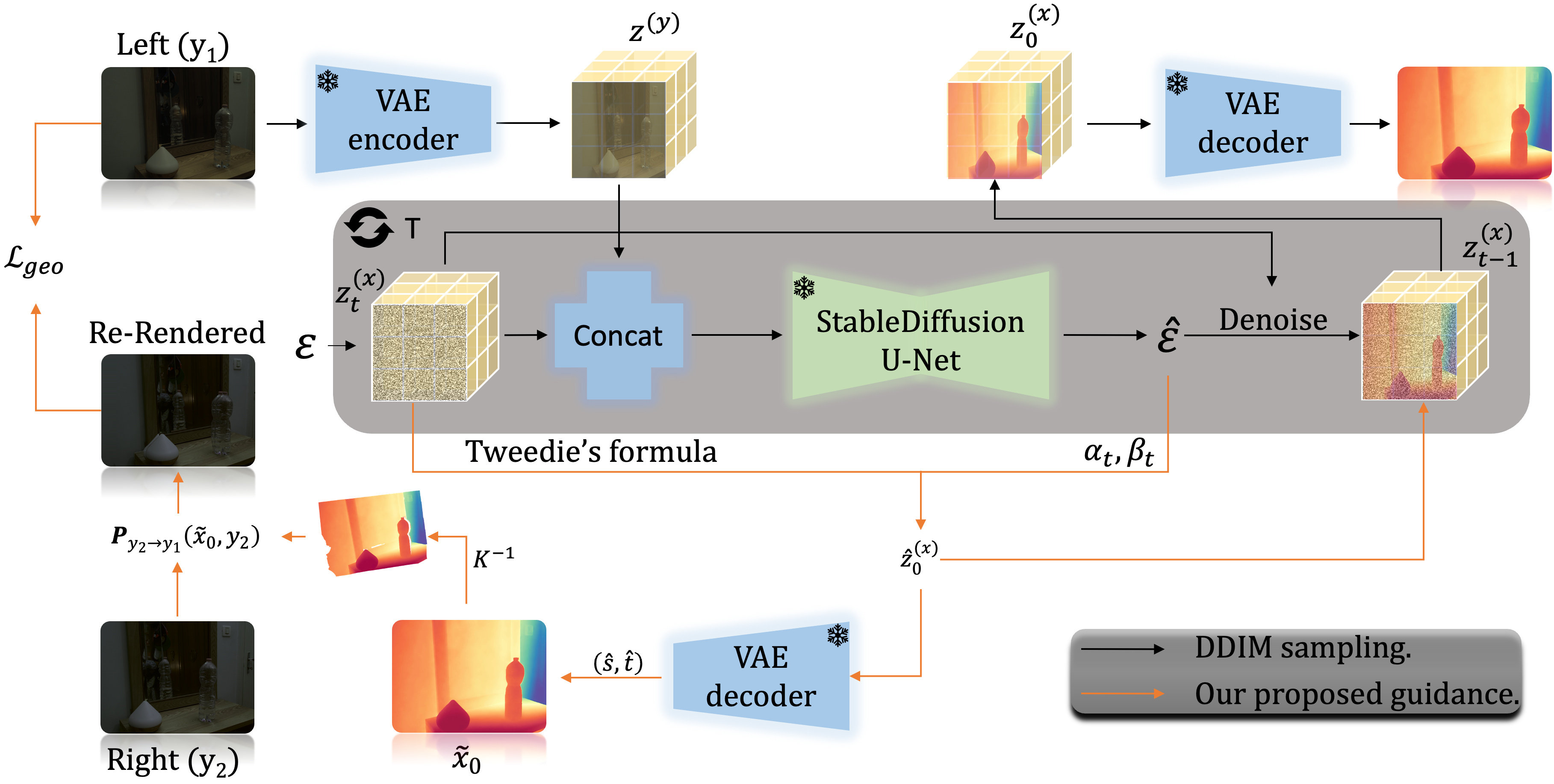}
  
  \end{center}
  \vspace{-1em}
  \caption{\textbf{Overview of the proposed framework.} GeoDiff is built upon DDIM~\cite{song2020denoising} sampling (black arrow) process with geometric guidance (yellow arrow), taking a stereo pair (or two views with known poses) as input and producing metric depth in meters for the left image. The process begins by encoding the left image through a VAE encoder and concatenating it with random noise to form the depth latent. During sampling, the one-step latent prediction $\hat{z}_0^{(x)}$ is computed using Tweedie's formula and decoded to pixel space. This prediction is then transformed to metric scale depth $\tilde{x}_0$ via learnable scale $\hat{s}$ and shift $\hat{t}$ parameters. A differentiable warping module $\bm{P}_{\bm{y}_2 \rightarrow \bm{y}_1}\left(\tilde{\bm{x}}_0, \bm{y}_2\right)$ leverages camera parameters to re-render the left image (detailed in Section~\ref{subsec:stereo-reproj} and Supplemental~\ref{subsec:supp_diff_warping}). The sampling process is guided by minimizing a geometric loss $\mathcal{L}_{geo}$ defined in Equation~\ref{eq:geo-loss}. }  
  \label{fig:framework}
  \vspace{-1.5em}
\end{figure*}

This section describes our proposed method in detail. Our method (see Figure~\ref{fig:framework}) aims to predict metric depth from a stereo pair (or two views with known poses). Specifically, given a stereo pair $\bm{y}_1, \bm{y}_2 \in \mathbb{R}^{H \times W \times 3}$ with known camera parameters, our goal is to estimate the metric depth map $\tilde{\bm{x}}$ corresponding to the first image $\bm{y}_1$. At a high level, our approach estimates metric depth by optimizing parameters that enable the reconstruction of one image from another through a depth map derived from a diffusion process.

\subsection{Metric depth parameterization}
\label{subsec:depth-param}
To estimate the metric depth $\tilde{\bm{x}}$ from the relative depth map $\bm{x}$ obtained from the DB-MDE model, we introduce a learnable linear transformation parameterized by scale and shift parameters. Specifically, we define the metric depth as:
\begin{equation} 
\label{eq:metric}
 \tilde{\bm{x}} = \text{softplus}(\hat{s}) \cdot \bm{x} + \text{softplus}(\hat{t}), 
\end{equation}
where $\hat{s}$ and $\hat{t}$ are the learnable scale and shift parameters, respectively. The softplus activation function $\text{softplus}(z) := \ln(1 + \exp(z))$, ensures that both the scale and shift are positive values, preventing negative depths.

\subsection{Geometric-Guided diffusion}
\label{subsec:geo-guided-diff}
Drawing inspiration from diffusion-based approaches to inverse problems \citep{song2021solving, chung2022improving, chung2022diffusion}, we formulate stereo depth estimation as an inverse problem (see Section~\ref{subsec:DM-Inverse}):
\begin{equation}
\label{eq:main_problem}
\bm{y}_2 = \bm{P}_{\bm{y}_1 \rightarrow \bm{y}_2} \left( \tilde{\bm{x}}, \bm{y}_1 \right) + \bm{e}, \bm{e} \sim \mathcal{N} (0, \sigma^2I)
\end{equation}
where $\bm{P}_{\bm{y}_1 \rightarrow \bm{y}_2}$ denotes the projection function mapping the metric depth $\tilde{\bm{x}}$ and the first image $\bm{y}_1$ to the second image $\bm{y}_2$, and $\bm{e}$ represents Gaussian noise. Our aim is to recover the metric depth map $\tilde{\bm{x}}$ using both $\bm{y}_1$ and the additional observation $\bm{y}_2$, leveraging the stereo geometry inherent in the forward model.

Following the derivation given in Equation~\ref{eq:cond_score_inv}, we can calculate the conditional score $\nabla_{\bm{z}_t} \log p \left(\bm{z}_t \mid \bm{y}_1, \bm{y}_2 \right)$ by:
\begin{align}
    \nabla_{\bm{z}_t} \log p & \left(\bm{z}_t \mid \bm{y}_1, \bm{y}_2 \right) \approx \nabla_{\bm{z}_t} \log p(\bm{z}_t \mid \bm{y}_1) \nonumber \\
    &- \lambda \nabla_{\bm{z}_t} \| \bm{y}_2 - \bm{P}_{\bm{y}_1 \rightarrow \bm{y}_2} (\tilde{\bm{x}}_0, \bm{y}_1) \|_2^2
\end{align}
where $\lambda$ is a tunable hyperparameter, $\nabla_{\bm{z}_t} \log p(\bm{z}_t \mid \bm{y}_1) \approx \bm{s}_\theta(\bm{z}_t, t, \bm{y}_1)$ is the pretrained score of the DB-MDE model and $\tilde{\bm{x}}_0$ is the one-step estimated metric depth at the current iteration, and can be computed using the one-step latent prediction $\hat{\bm{z}}_0$:
\begin{equation}
    \label{eq:metric}
    \tilde{\bm{x}}_0 = \text{softplus}(\hat{s}) \bm{D}(\hat{\bm{z}}_0) + \text{softplus}(\hat{t})
\end{equation}
A detailed derivation can be found in our Supplemental~\ref{subsec:supp_conditional_score}. 

However, computing the forward warping function $\bm{P}_{\bm{y}_1 \rightarrow \bm{y}_2}$ is not preferred due to holes and computational complexity (see Supplemental~\ref{subsec:supp_diff_warping}). Therefore, we adopt a backward warping function $\bm{P}_{\bm{y}_2 \rightarrow \bm{y}_1}$ instead of a forward warping. We also follow previous works \cite{godard2017unsupervised, godard2019digging} to employ a linear combination of SSIM and L1 losses to calculate the geometric reprojection loss:
\begin{align}
\label{eq:geo-loss}
\mathcal{L}_{geo}  &= \eta (1 - \text{SSIM}\left(\bm{y}_1, \bm{P}_{\bm{y}_2 \rightarrow \bm{y}_1}\left(\tilde{\bm{x}}_0, \bm{y}_2\right)\right)) / 2 \nonumber \\
&+ (1 - \eta) \|\bm{y}_1 - \bm{P}_{\bm{y}_2 \rightarrow \bm{y}_1}\left(\tilde{\bm{x}}_0, \bm{y}_2\right)\|_1
\end{align}
where $\eta$ is the hyperparameter that balances the two losses. Our final score function is:
\begin{align}
\label{eq:final_score}
\nabla_{\bm{z}_t} \log p(\bm{z}_t &\mid \bm{y}_1, \bm{y}_2) = \bm{s}_\theta(\bm{z}_t, t, \bm{y}_1) \nonumber \\
&- \lambda \nabla_{\bm{z}_t} \mathcal{L}_{geo} \left(\bm{y}_1, \bm{P}_{\bm{y}_2 \rightarrow \bm{y}_1}\left(\tilde{\bm{x}}_0, \bm{y}_2\right)\right)
\end{align}
At every diffusion sampling step, we leverage this score function to update the current latent $\bm{z}_t$, while simultaneously updating the scale $\hat{s}$ and shift $\hat{t}$ using the reprojection gradient. The detailed algorithm is shown in the Supplementary~\ref{sec:supp_algorithm}. 
\paragraph{Generalization.} Acquisition of stereo image pairs in unconstrained real-world scenarios presents significant practical challenges, typically necessitating a calibrated dual-camera setup with precise side-by-side alignment. Our proposed framework, however, extends beyond traditional stereo configurations to accommodate arbitrary two-view settings with known relative poses, substantially enhancing its applicability across diverse deployment contexts. Our method only requires relative transformation for performing differentiable warping operations (detailed in Supplemental~\ref{subsec:supp_diff_warping}). For in-the-wild image pairs lacking calibration metadata, we leverage recent advances in foundation models for dense 3D reconstruction~\cite{wang2024dust3r} to estimate the requisite camera intrinsic and extrinsic parameters.
\paragraph{Regularization.} To improve the stability of our optimization, we introduce a global scale hyperparameter $g_s$ and regularization loss for scale $\hat{s}$ and shift $\hat{t}$. Through our experiments, we observe that each scene in the wild has different depth scale. The further the depth is, the better the warped image can be rendered, thus minimizing the $\mathcal{L}_{geo}$, and keep enforcing the depth further away. It leads to the wrong sampling trajectory of the diffusion model. Therefore, we opt to pre-select a global scale $g_s$ and also apply $L_2$ regularize on the $\hat{s}$ and shift $\hat{t}$ parameters. Now, the $\tilde{x}_0$ and the total optimization loss $\mathcal{L}_{geo}$ become:
\begin{align}
    &\tilde{\bm{x}}_0 = g_s [\text{softplus}(\hat{s}) \cdot \hat{\bm{x}}_0 + \text{softplus}(\hat{t})] \\
    &\mathcal{L}_{geo} = \mathcal{L}_{geo} + \gamma (||\hat{s}||^2_2 + ||\hat{t}||^2_2)
\end{align}
where we set $\gamma:= 1e-2$ for all of our experiments. We provide in-depth discussion regarding regularization and global scale $g_s$ in the Supplementary~\ref{subsec:supp_reg}.
\paragraph{Discussion.} Unlike prior works that directly utilize the reprojection loss for self-supervised depth estimation \cite{godard2017unsupervised, godard2019digging} or online adaptation \cite{tonioni2019learning, zhang2019online}, our method incorporates the gradient of this loss as additional guidance within the diffusion sampling process. As Section~\ref{subsec:ablation} demonstrates, optimizing this reprojection loss in the raw pixel space is highly susceptible to noise and can lead to inferior depth maps. Consequently, previous methods have employed smoothness regularizers to mitigate this issue. By embedding the loss into the diffusion sampling framework, our diffusion model inherently acts as an implicit regularizer, effectively stabilizing the optimization and obviating the need for explicit smoothness constraints.

\section{Experiments}
\label{sec:experiments}
In this section, we first describe our experimental settings at Section~\ref{subsec:exp-setup}. Then, we showcase our experiment results in Section~\ref{subsec:exp-res}. Finally, we perform ablation study at Section~\ref{subsec:ablation}.
\begin{figure*}[t]
  \begin{center}

  \includegraphics[width=\textwidth]{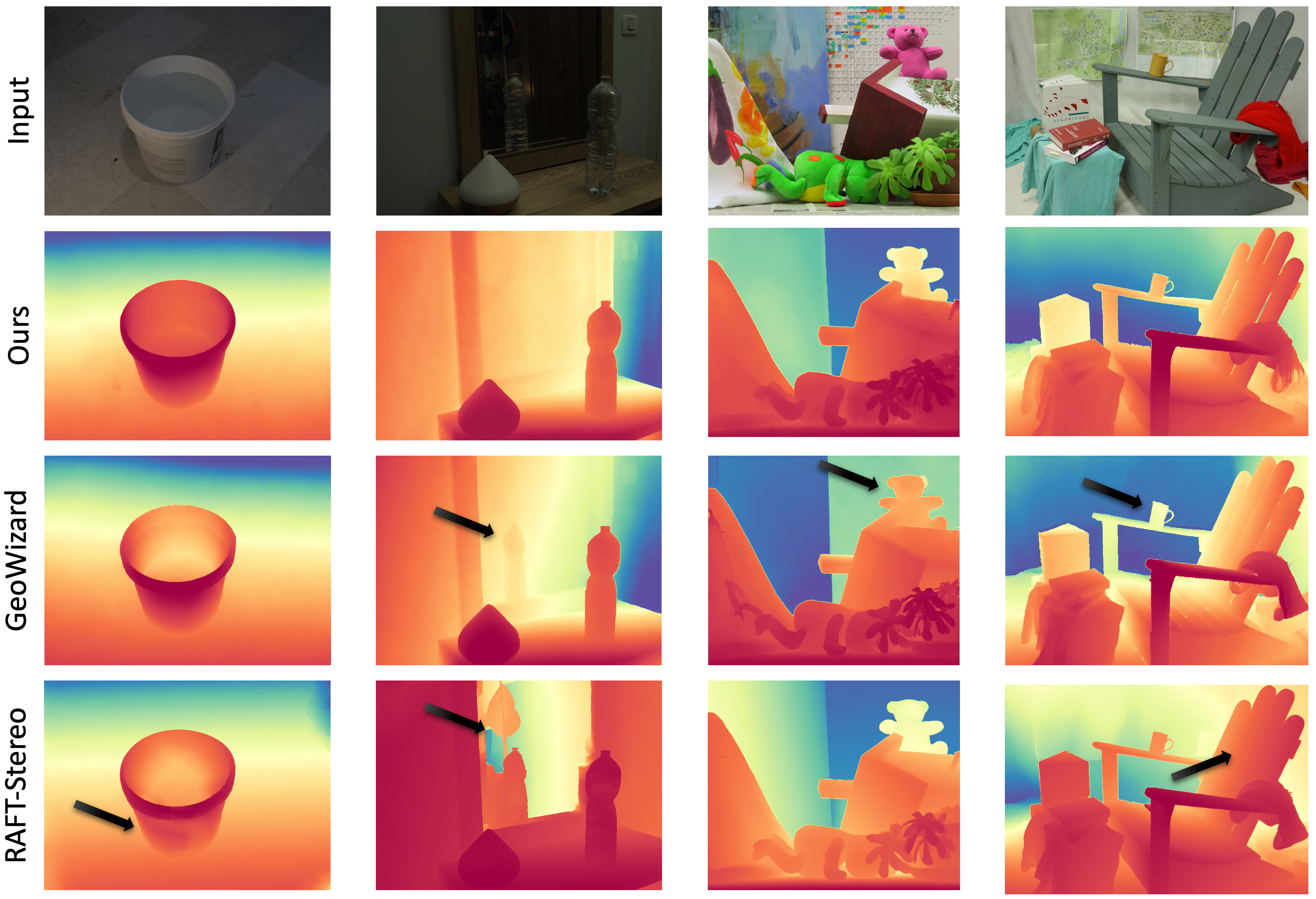}
  \end{center}
  \vspace{-1em}
  \caption{\textbf{Qualitative results.} Our method showcases superior depth prediction compared to competing methods. For instance, in the challenging Mirror scene (second column), our model accurately predicts the reflective surface, outperforming both GeoWizard~\cite{fu2024geowizard} and RAFT-Stereo, which struggle in such cases. Additionally, our approach preserves finer details (last column), showcasing the effectiveness of our proposed geometric guidance.}
  \label{fig:qualitative}
  \vspace{-1.5em}
\end{figure*}

\subsection{Experiment setup}
\label{subsec:exp-setup}

\paragraph{Implementation details.} Our method is built upon Marigold~\cite{ke2024repurposing}. Specifically, we employ their public pretrained model and modify it with our guidance. While doing optimization, we completely freeze the trained weights of the model. The learning rate for optimizing parameters in Section~\ref{subsec:depth-param} and the depth latents are set to $1e-2$. Following Marigold~\cite{ke2024repurposing}, we use an ensemble of 10 depth samples as our final prediction for computing all metrics. For fair comparison, all diffusion-based methods listed in Table~\ref{tab:depth_comparison} are also results of ensemble prediction. All experiments are conducted on a single NVIDIA RTX A6000 GPU. 
\paragraph{Dataset.} We perform zero-shot evaluation of our method on three datasets: KITTI-2015 \citep{menze2015joint}, Booster \citep{zamaramirez2024booster} and Middlebury \citep{scharstein2014high}. While KITTI-2015 and Middlebury are the two common outdoor and indoor benchmark for depth estimation, Booster is a more recent depth benchmark focusing on specular and transparent objects. We also sample a subset of multi-view depth dataset Tanks and Temples for evaluation of arbitrary two-view setting. More detail is shown at Supplemental~\ref{sec:supp_dataset_detail}.
\paragraph{Evaluation metrics.} Following previous works~\cite{ke2024repurposing, fu2024geowizard}, we conduct zero-shot metric depth estimation by measuring three metrics including mean absolute error (AbsRel), $\delta1$ accuracy. Additionally, we measure the root mean square error (RMSE)  $\frac{1}{M} \sum_{i=1}^M ||d^{gt}_i - d^{pred}_i ||^2_2$, where $M$ denotes the number of samples, and $d^{gt} , d^{pred}$ represent ground truth depth and predicted depth. To comprehensively demonstrate our method's effectiveness, we report results for both predicted affine-invariant depth map and our metric depth maps. For affine-invariant depth map, we use the common protocol employed in previous works~\cite{ke2024repurposing, fu2024geowizard} that align predicted depth map to ground truth via least-squares fitting. Conversely, our metric depth map is directly compare with ground truth depth without any alignment.
\paragraph{Competing methods.} Our method utilizes a stereo pair as input while leveraging pretrained monocular priors without any stereo-specific training. To the best of our knowledge, this represents the first training-free approach that adapts monocular priors for stereo depth estimation. Consequently, there are no direct competing methods in the same setting. We therefore conduct comprehensive comparisons across related domains to contextualize our contributions. Since our method is based on monocular priors, we compare against recent state-of-the-art affine-invariant MDE methods including Marigold~\cite{ke2024repurposing}, GeoWizard~\cite{fu2024geowizard}, and MiDas~\cite{birkl2023midas}. Following their evaluation protocol, we align their predicted depths with ground truth using least squares fitting~\cite{ranftl2020towards, ke2024repurposing, fu2024geowizard}. As our method also recovers metric depth, we compare with leading metric depth estimation approaches: ZoeDepth~\cite{bhat2023zoedepth}, UniDepth~\cite{piccinelli2024unidepth}, and DepthPro~\cite{bochkovskii2024depth}. In this setting, we directly compare raw metric depth outputs without any alignment. Given our use of stereo pairs as input, we additionally benchmark against stereo matching-based methods including RAFT-Stereo~\cite{lipson2021raft} and IGEV-Stereo~\cite{xu2023iterative}. While these methods primarily predict disparity rather than depth, the comparison remains relevant as disparity ($disp$) can be directly converted to depth ($d$) given the baseline ($b$) and camera focal length ($f$): $d=bf/disp$.
\subsection{Experimental results}
\label{subsec:exp-res}

\begin{table*}[!ht]
\scriptsize
\centering
\caption{\textbf{Quantitative comparison with monocular depth methods on KITTI-Stereo, Booster, and Middlebury Datasets.} Our method outperforms existing approaches in both aligned and non-aligned settings. ``GT-Aligned" indicates whether predictions are affine-invariant and require alignment with ground truth. Color coding: \colorbox{red!50}{best}, \colorbox{orange!50}{second-best}, and \colorbox{yellow!50}{third-best} results. For metrics, $\downarrow$ indicates lower is better while $\uparrow$ indicates higher is better. We denote $^\dagger$ as our result for metric depth, and the other result is affine-invariant depth.}
\label{tab:depth_comparison}
\resizebox{1\textwidth}{!}{
\begin{tabular}{l c ccc ccc ccc}
\toprule
\multirow{2}{*}{\textbf{Method}} & \multirow{2}{*}{\textbf{GT-Aligned}} & \multicolumn{3}{c}{\textbf{KITTI-Stereo}} & \multicolumn{3}{c}{\textbf{Booster}} & \multicolumn{3}{c}{\textbf{Middlebury}} \\
\cmidrule(lr){3-5} \cmidrule(lr){6-8} \cmidrule(lr){9-11}
 & & \textbf{AbsRel}$\downarrow$ & $\bm{\delta 1}$$\uparrow$ & \textbf{RMSE}$\downarrow$ & \textbf{AbsRel}$\downarrow$ & $\bm{\delta 1}$$\uparrow$ & \textbf{RMSE}$\downarrow$ & \textbf{AbsRel}$\downarrow$ & $\bm{\delta 1}$$\uparrow$ & \textbf{RMSE}$\downarrow$ \\
\midrule
MiDas~\cite{birkl2023midas}  & {\color{green}\checkmark} & 0.63 & 0.24 & 11.72 & \best{3}{0.18} & 0.71 & 0.23 & 0.22 & 0.72 & 2.09 \\
Marigold~\cite{ke2024repurposing}  & {\color{green}\checkmark} & \best{3}{0.13} & 0.85 & 4.81 & \best{1}{0.04} & \best{1}{0.98} & \best{1}{0.06} & \best{2}{0.14} & \best{3}{0.83} & \best{3}{1.61} \\
GeoWizard~\cite{fu2024geowizard}  & {\color{green}\checkmark} & 0.18 & 0.75 & 5.7 & \best{1}{0.04} & \best{2}{0.96} & \best{2}{0.08} & \best{3}{0.15} & 0.81 & \best{2}{1.55} \\
\midrule
ZoeDepth~\cite{bhat2023zoedepth} & {\color{red}\texttimes} & 0.74 & $\backslash$ & 16.26 & 7.37 & $\backslash$ & 7.72 & 0.60 & $\backslash$ & 6.04 \\
UniDepth~\cite{piccinelli2024unidepth} & {\color{red}\texttimes} & 0.19 & \best{2}{0.86} & \best{3}{4.18} & 4.55 & $\backslash$ & 4.91 & 0.61 & $\backslash$ & 6.03 \\
DepthPro~\cite{bochkovskii2024depth} & {\color{red}\texttimes} & 0.16 & 0.81 & 4.43 & 0.34 & 0.51 & 0.67 & 0.54 & 0.02 & 5.60 \\
\midrule
Ours$^\dagger$ & {\color{red}\texttimes} & \best{1}{0.07} & \best{1}{0.91} & \best{2}{3.94} & \best{2}{0.11} & \best{3}{0.81} & \best{3}{0.18} & \best{1}{0.11} & \best{2}{0.85} & 2.31 \\
Ours & {\color{green}\checkmark} & \best{2}{0.09} & \best{1}{0.91} & \best{1}{3.72} & \best{1}{0.04} & \best{1}{0.98} & \best{1}{0.06} & \best{1}{0.11} & \best{1}{0.87} & \best{1}{1.41} \\
\bottomrule
\end{tabular}
}
\end{table*}

\subsubsection{Comparison with Monocular Depth Estimation.}
\label{subsubsec:compare-mono}
\paragraph{Quantitative results.} We present our quantitative results in Table~\ref{tab:depth_comparison}. Our method demonstrates superior performance in both aligned and non-aligned settings across all datasets. In the depth affine-invariant setting with ground truth alignment, our method significantly outperforms existing approaches across all metrics. We consistently surpass our baseline method Marigold~\cite{ke2024repurposing} by a substantial margin. On both KITTI-Stereo and Middlebury datasets, our affine-invariant depth outperforms all competing methods including MiDas~\cite{birkl2023midas}, Marigold~\cite{ke2024repurposing}, and GeoWizard~\cite{fu2024geowizard}. In the non-aligned setting, our method also exhibits competitive performance across all datasets, even when compared with affine-invariant methods that require ground truth alignment. On KITTI-Stereo (an outdoor dataset), we achieve the lowest AbsRel of $0.07$ and highest $\delta1$ of $0.91$. On Middlebury (an indoor dataset), our approach outperforms in AbsRel ($0.11$) and $\delta1$ ($0.85$) metrics, falling short only in the RMSE metric. For Booster, a challenging dataset with non-Lambertian surfaces including transparent and reflective objects, we consistently outperform metric depth estimation methods such as ZoeDepth~\cite{bhat2023zoedepth}, UniDepth~\cite{piccinelli2024unidepth}, and DepthPro~\cite{bochkovskii2024depth}. This highlights our method's robustness when handling challenging surface properties. It is notable that without retraining Marigold~\cite{ke2024repurposing}, our method outperforms diffusion-based monocular baselines in non-aligned settings while achieving superior performance with alignment. These results demonstrate our approach's ability to produce accurate metric depth maps without requiring ground truth alignment, while still excelling when alignment is applied.

\paragraph{Qualitative results.} We present our qualitative results in Figure~\ref{fig:demo} and Figure~\ref{fig:qualitative}. Our method, which leverages both strong geometric guidance and pretrained diffusion priors, effectively captures fine-grained details while accurately representing transparent and specular objects. When compared to our baseline Marigold~\cite{ke2024repurposing}, our approach eliminates several erroneous artifacts due to our geometric guidance (as shown in Figure~\ref{fig:demo}). In comparison with GeoWizard~\cite{fu2024geowizard}, our method correctly handles reflective surfaces such as mirrors and captures more detailed depth information in indoor scenes, as illustrated in Figure~\ref{fig:qualitative}. Although DepthPro~\cite{bochkovskii2024depth} may look sharper in 2D visual results, our depth predictions are superior in terms of metric accuracy, as depth inherently represents 3D information. More visualization results are presented in Supplemental~\ref{sec:supp_qualitative}.

\subsubsection{Comparison with Stereo Depth Estimation}
\begin{table}
\scriptsize
\centering
\caption{\textbf{Quantitative comparison with stereo depth on Booster Dataset.} Our methods performs comparably or better than stereo methods despite no explicit stereo training. While both aligned and non-aligned results are reported, we emphasize raw metric predictions (non-aligned), with affine-invariant results provided only for reference.}
\label{tab:stereo_comparison}

\begin{tabular}{l c ccc}
\toprule
\multirow{2}{*}{\textbf{Method}} & \multirow{2}{*}{\textbf{GT-Aligned}} & \multicolumn{3}{c}{\textbf{Booster}} \\
\cmidrule(lr){3-5}
 & & \textbf{AbsRel}$\downarrow$ & $\bm{\delta 1}$$\uparrow$ & \textbf{RMSE}$\downarrow$ \\
\midrule
RAFT-Stereo~\cite{lipson2021raft} & {\color{red}\texttimes} & 5.4 & $\backslash$ & 18.89 \\
IGEV-Stereo~\cite{xu2023iterative} & {\color{red}\texttimes} & \cellcolor{red!50}0.10 & \cellcolor{red!50}0.94 & \cellcolor{orange!50}0.31 \\
\midrule
Ours$^\dagger$ & {\color{red}\texttimes} & \cellcolor{orange!50}0.11 & \cellcolor{orange!50}0.81 & \cellcolor{red!50}0.18 \\
Ours & {\color{green}\checkmark} & \cellcolor{gray!50}0.04 & \cellcolor{gray!50}0.98 & \cellcolor{gray!50}0.06 \\
\bottomrule
\end{tabular}
\end{table}
\paragraph{Quantitative results.} We present our quantitative result in stereo depth estimation setting at Table~\ref{tab:stereo_comparison}. It is worth noting that previous stereo methods such as RAFT-Stereo~\cite{lipson2021raft} and IGEV-Stereo~\cite{xu2023iterative} have been trained KITTI-Stereo dataset and are extensively tuned for Middlebury, yet remain unevaluated on the challenging Booster dataset. Therefore, for fair comparison with our method--as the pretrained never encountered Booster during training--we conduct comparative analysis on this dataset. In non-alignment setting, we achieve on-par result compare to strong stereo method IGEV-Stereo in AbsRel and $\delta1$ ($0.11$ and $0.81$, respectively). Notably, we outperform both stereo methods in the RMSE metric with a value of $0.18$. These results demonstrate that our geometry-guided diffusion prior approach performs consistently on par or better with stereo method that has been exposed to stereo training data.  
\paragraph{Qualitative results.} As shown in Figure~\ref{fig:demo} and Figure~\ref{fig:qualitative}, our method excels on cases that contain non-Lambertian surfaces such as transparent materials and reflective surfaces. We speculate that though stereo methods are extensively trained on stereo data, they heavily rely on cost volume-based approach, which inherently struggles with transparent or textureless surfaces~\cite{zamaramirez2024booster}. Our method, on the other hand, inherits strong monocular priors from diffusion models guided with explicit geometry guidance during optimization, thus addressing the cost volume limitation. Furthermore, a strong geometry-guided monocular prior also help to achieve shaper depth in fine-grain areas (see Middlebury results in Figure~\ref{fig:qualitative}).  
\subsubsection{Arbitrary two-view settings}
\label{subsubsec:exp-two-view}
\begin{table}
\scriptsize
\centering
\caption{\textbf{Quantitative comparison with stereo depth on Tank and Temple.} Our method excels in metric depth estimation, and on par with Dust3r when aligned with ground truth. }
\label{tab:two_views}

\begin{tabular}{l c ccc}
\toprule
\multirow{2}{*}{\textbf{Method}} & \multirow{2}{*}{\textbf{GT-Aligned}} & \multicolumn{3}{c}{\textbf{Tanks and Temples}} \\
\cmidrule(lr){3-5}
 & & \textbf{AbsRel}$\downarrow$ & $\bm{\delta 1}$$\uparrow$ & \textbf{RMSE}$\downarrow$ \\
\midrule
Dust3r~\cite{wang2024dust3r} & {\color{red}\texttimes} & 0.51 & 0.35 & \best{1}{0.56} \\
Ours$^\dagger$ & {\color{red}\texttimes} & \best{1}{0.47} & \best{1}{0.38} & 0.62 \\
\midrule
Dust3r~\cite{wang2024dust3r} & {\color{green}\checkmark} & \best{1}{0.07} & \best{1}{0.93} & \best{1}{0.17} \\
Ours & {\color{green}\checkmark} & 0.08 & 0.92 & 0.18 \\
\bottomrule
\end{tabular}
\end{table}
\begin{figure}
    \centering
    \includegraphics[width=\linewidth]{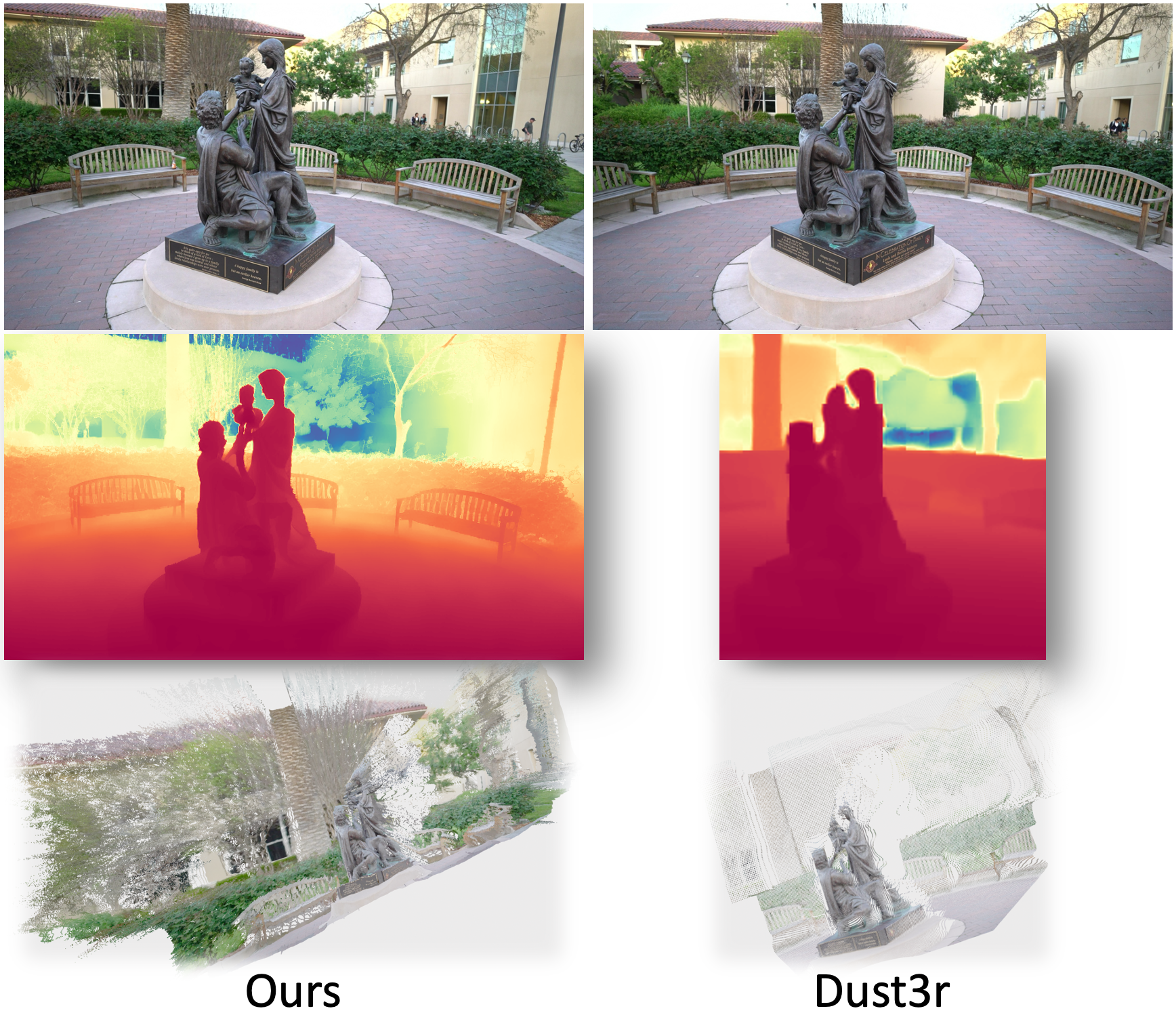}
    \caption{From top to bottom. \textbf{Top:} Two known-pose arbitrary images. \textbf{Middle:} Depth map predictions of our method and Dust3r~\cite{wang2024dust3r}. \textbf{Bottom:} Our point cloud and Dust3r point cloud.}
    \label{fig:exp-two-view}
\end{figure}
As established in Section~\ref{subsec:geo-guided-diff}, our methodology extends beyond traditional stereo setups to general two-view configurations with known camera poses. While numerous multi-view stereo techniques can infer depth from multiple viewpoints~\cite{wu2024gomvs, yao2018mvsnet, xu2024robustmvs}, our work focuses specifically on the two-view paradigm. We compare our method with Dust3r~\cite{wang2024dust3r}, which is trained on two views to generate a point map from a single view. Quantitative results are presented in Table~\ref{tab:two_views}, with qualitative evaluation shown in Figure~\ref{fig:exp-two-view}. Compared to Dust3r~\cite{wang2024dust3r}, our approach produces significantly more detailed depth maps with enhanced structural fidelity. Although Dust3r generates visually plausible point clouds, their reconstructions are limited to an unknown scale factor. In contrast, our method produces metric point clouds that accurately represent scene geometry at absolute scale, enabling precise spatial measurements and supporting reliable downstream applications.

\subsection{Ablation studies}
\label{subsec:ablation}
\subsubsection{Non-optimality of the reprojection loss}
\label{sec:ab1} 
\begin{figure}
    \centering
    \includegraphics[width=\linewidth]{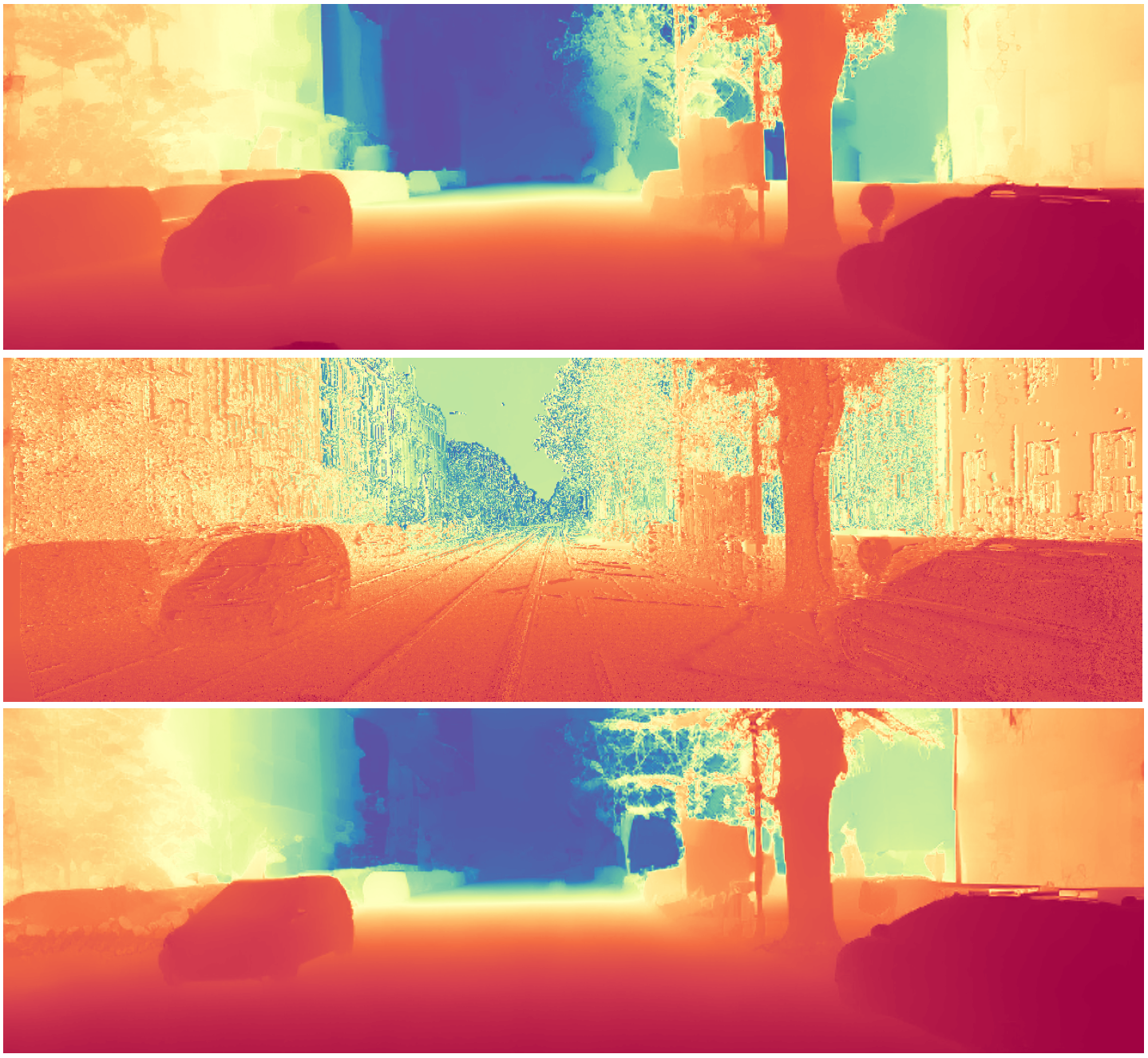}
    \caption{\textbf{Ablation.} From top to bottom: \textit{Marigold prediction}, \textit{optimization with reprojection loss only}, and \textit{ours combining diffusion with reprojection loss guidance}. Reprojection loss optimization leads to noisy and suboptimal depth, while using it as the guidance for diffusion model helps improve the results.}
    \label{fig:ab1}
\end{figure}

We empirically investigated the noise sensitivity of reprojection loss discussed in Section~\ref{sec:method}. By directly optimizing metric depth without the diffusion framework—initializing with Marigold outputs and optimizing both depth and scale/shift parameters over 50 iterations—we observe significantly noisy results as shown in Figure \ref{fig:ab1}. In contrast, our diffusion-based approach preserves fine details while avoiding noise artifacts. We hypothesize that the diffusion sampling process inherently regularizes the optimization toward the true depth distribution, eliminating the need for explicit smoothness regularizers \citep{godard2017unsupervised, godard2019digging}.

\subsubsection{Effectiveness of the reprojection loss}
To investigate whether reprojection loss primarily enhances depth quality or merely calibrates scale/shift parameters, we conducted an ablation study restricting this loss to optimize only scale and shift without influencing diffusion latents. As shown in Table \ref{tab:ablation_study}, experiments on KITTI-Stereo demonstrate significant performance degradation under this configuration, confirming the crucial role of reprojection loss guidance in refining the diffusion process for accurate metric depth estimation.

\begin{table}
\centering
\caption{Ablation Study on KITTI-Stereo Dataset}
\label{tab:ablation_study}
\begin{tabular}{lcc}
\toprule
Method & \textbf{AbsRel}$\downarrow$ & $\bm{\delta 1} \uparrow$ \\
\midrule
Learning scale and shift only & 0.14 & 0.80 \\
Our full model & 0.07 & 0.91 \\
\bottomrule
\end{tabular}
\end{table}

\section{Conclusion and Limitation}
\label{sec:conclusion}
In this work, we introduced a novel framework that extends diffusion-based monocular depth estimation (DB-MDE) models to metric depth prediction by incorporating stereo settings and an inverse problem (IP) approach. By leveraging pretrained latent diffusion models (LDMs) with stereo geometric guidance, our method effectively addresses scale and shift ambiguities inherent in monocular depth estimation. Extensive experiments demonstrate its robustness across diverse environments, including indoor, outdoor, and challenging specular scenes, all without requiring domain-specific retraining.

Despite its strengths, our approach has certain limitations. First, it relies on pretrained monocular depth estimation models, meaning the quality of depth predictions is dependent on the strength of the prior. A more expressive or robust MDE model could further enhance performance. Second, like other DB-MDE approaches, our method incurs slow inference times due to the iterative nature of diffusion-based sampling. Future work could explore accelerated sampling techniques or lighter-weight diffusion architectures to improve efficiency while maintaining accuracy.


\clearpage
\setcounter{page}{1}
\maketitlesupplementary

In this supplementary material, we first provide additional derivations and insights in Section~\ref{sec:supp_detail}. We then present our method through pseudocode in Section~\ref{sec:supp_algorithm}, followed by implementation details in Section~\ref{sec:supp_implementation_detail}. Dataset specifications are described in Section~\ref{sec:supp_dataset_detail}, while limitations and future work are discussed in Section~\ref{sec:supp_limitation}. Finally, additional experimental results are presented in Section~\ref{sec:supp_qualitative}.

\section{Detailed method}
\label{sec:supp_detail}
\subsection{Conditional Score Derivation}
\label{subsec:supp_conditional_score}
In this section, we provide the complete derivation of our conditional score. Applying Bayes' theorem, the score function of the conditional distribution can be expressed as:
\begin{align}
\nabla_{\bm{z}_t} \log p(\bm{z}_t \mid \bm{y}_1, \bm{y}_2) &= \nabla_{\bm{z}_t} \log p(\bm{z}_t \mid \bm{y}_1) \nonumber \\
&+ \nabla_{\bm{z}_t} \log p(\bm{y}_2 \mid \bm{z}_t, \bm{y}_1)
\end{align}
Under mild assumptions \citep{chung2022diffusion}, and a decoder $\bm{D}$ that maps the latent back to the image space, we can approximate this score function using:
\begin{align}
\nabla_{\bm{z}_t} \log p(\bm{y}_2 \mid \bm{z}_t, \bm{y}_1) &\approx \nabla_{\bm{z}_t} \log p(\bm{y}_2 \mid \hat{\bm{z}}_0, \bm{y}_1) \nonumber \\
&\approx \nabla_{\bm{z}_t} \log p(\bm{y}_2 \mid \bm{D}(\hat{\bm{z}}_0), \bm{y}_1)
\end{align}
where $\hat{\bm{z}}_0$ is estimated using Tweedie's formula \citep{robbins1992empirical}: 
\begin{equation}
\label{eq:tweedie}
\hat{\bm{z}}_0=\frac{1}{\sqrt{\bar{\alpha}_t}}\left(\bm{z}_t+\sqrt{1-\bar{\alpha}_t} \bm{s}_\theta\left(\bm{z}_t, t, \bm{y}_1 \right)\right)
\end{equation}
Leveraging the Gaussian noise model assumption in Equation \ref{eq:main_problem}, we get:
\begin{equation}
\nabla_{\bm{z}_t} \log p\left(\bm{y}_2 \mid \bm{z}_t, \bm{y}_1\right) \simeq-\frac{1}{\sigma^2} \nabla_{\bm{z}_t}\left\|\bm{y}_2 -\bm{P}_{\bm{y}_1 \rightarrow \bm{y}_2}\left(\tilde{\bm{x}}_0, \bm{y}_1\right)\right\|_2^2
\end{equation}
in which $\tilde{\bm{x}}_0$ is the metric depth calculated following the Equation \ref{eq:metric}.

Thus, the final conditional score function is:
\begin{align}
    \nabla_{\bm{z}_t} \log p \left(\bm{z}_t \mid \bm{y}_1, \bm{y}_2 \right) &\approx \nabla_{\bm{z}_t} \log p(\bm{z}_t \mid \bm{y}_1) \nonumber \\
    &- \lambda \nabla_{\bm{z}_t} \| \bm{y}_2 - \bm{P}_{\bm{y}_1 \rightarrow \bm{y}_2} (\tilde{\bm{x}}_0, \bm{y}_1) \|_2^2
\end{align}

\subsection{Differentiable warping}
\label{subsec:supp_diff_warping}
As discussed in Section~\ref{subsec:stereo-reproj} and Section~\ref{subsec:geo-guided-diff}, we use differentiable warping to render novel view given predicted depth. Then, our method leverages given RGB input image to calculate photometric loss as guidance (see Equation~\ref{eq:geo-loss}) for diffusion process. There are two design choices for warping operator, which are forward warping operator $\bm{P}_{\bm{y}_1 \rightarrow \bm{y}_2} \left( \bm{x}_1, \bm{y}_1 \right)$, which projects $\bm{y}_1$ onto $\bm{y}_2$; and the backward warping operator $\bm{P}_{\bm{y}_2 \rightarrow \bm{y}_1}\left(\bm{x}_1, \bm{y}_2\right)$. If one opts to use forward warping as renderer, $\mathcal{L}_{geo}$ in Equation~\ref{eq:geo-loss} has the following form:
\begin{align}
\label{eq:forward-geo-loss}
\mathcal{L}_{geo}^{forward}  &= \eta (1 - \text{SSIM}\left(\bm{y}_2, \bm{P}_{\bm{y}_1 \rightarrow \bm{y}_2}\left(\tilde{\bm{x}}_0, \bm{y}_1\right)\right)) / 2 \nonumber \\
&+ (1 - \eta) \|\bm{y}_2 - \bm{P}_{\bm{y}_1 \rightarrow \bm{y}_2}\left(\tilde{\bm{x}}_0, \bm{y}_1\right)\|_1
\end{align}

otherwise, the backward warping could also be used with the form:
\begin{align}
\label{eq:backward-geo-loss}
\mathcal{L}_{geo}^{backward}  &= \eta (1 - \text{SSIM}\left(\bm{y}_1, \bm{P}_{\bm{y}_2 \rightarrow \bm{y}_1}\left(\tilde{\bm{x}}_0, \bm{y}_2\right)\right)) / 2 \nonumber \\
&+ (1 - \eta) \|\bm{y}_1 - \bm{P}_{\bm{y}_2 \rightarrow \bm{y}_1}\left(\tilde{\bm{x}}_0, \bm{y}_2\right)\|_1
\end{align}

Now, we discuss the design choice of the two options. Given a source image ($\bm{y}_1$), target image ($\bm{y}_2$), intrinsic camera matrices $K_1, K_2$, and camera-to-world extrinsic matrices $E_1, E_2$ for the source and target views respectively, we establish a generalized framework for our method. We explicitly represent both camera extrinsics to handle arbitrary camera configurations rather than just using a single relative transformation $T_{\bm{1} \rightarrow \bm{2}}$ as in Equation~\ref{eq:forward-warp}. In the specific case of calibrated stereo pairs captured simultaneously by a binocular rig, the transformation simplifies to a pure translation. However, for arbitrarily captured image pairs, the complete extrinsic matrices are necessary to accurately transform points between the two coordinate systems.
\paragraph{Forward warping} maps pixels from a source image ($\bm{y}_1$) to positions in target image ($\bm{y}_2$). The target coordinate is formulated as:
\begin{equation}
\label{eq:forward-warp_supp}
\bm{c}_2 \thicksim K_2  E_2^{-1}  E_1  \bm{x}_1(\bm{c}_1)  K_1^{-1}  \bm{c}_1
\end{equation}

where $\bm{c}_1$ and $\bm{c}_2$ denote the homogeneous pixel coordinates in $\bm{y}_1$ and $\bm{y}_2$, respectively, and $\bm{x}_1(\bm{c}_1)$ represents the depth at pixel $\bm{c}_1$ in $\bm{y}_1$. After getting the corresponding pixel coordinates, we can "splat" each source pixel to its corresponding location in target view. However, there are a few implementation challenges. A fundamental issue is that some target pixels might not receive any values, creating holes in the warped image. These voids occur due to disocclusions (regions visible in the target view but occluded in the source view) and sampling disparities (discrete source pixels mapping to non-integer target coordinates with gaps between them). Addressing these holes requires complex post-processing techniques such as depth-aware inpainting or multi-scale filtering. The non-integer mapping of source pixels to target coordinates further introduces discretization errors and potential aliasing, requiring appropriate interpolation strategies. Depth map inaccuracies are particularly problematic at discontinuities, where slight errors can significantly distort the warped result, making it even more difficult to apply to our framework. From a computational perspective, the unpredictable memory access patterns inherent in forward warping present optimization difficulties, particularly for parallel processing implementations. 

\paragraph{Backward warping} pulls back pixels from target image ($\bm{y}_2$) to source image ($\bm{y}_1$). It is worth noting that our backward warping is different from previous works~\cite{lichy2024nvtorchcam, godard2017unsupervised}, where they perform backward warping from source to target given target depth. One the other hand, we warp from target back to source using source depth. Specifically, our backward warping is formulated as:
\begin{equation}
\label{eq:back-warp_supp}
\bm{c}_2 \thicksim K_2 E_2 E_1^{-1} \bm{x}_1(\bm{c}_1)  K_1^{-1} \bm{c}_1
\end{equation}
After computing the corresponding pixel coordinates, we sample pixel colors from the target image at these new coordinates. Since these coordinates are generally non-integer, we employ bilinear interpolation for color sampling. This approach inherently avoids the hole artifacts characteristic of forward warping methods. For this reason, we adopt backward warping as our rendering technique throughout this work.

\paragraph{Discussion.} We explored several alternative techniques that ultimately proved suboptimal. Initial experiments with point cloud rasterization from Pytorch3D~\cite{ravi2020pytorch3d} revealed high sensitivity to point diameter and opacity parameters, resulting in rendering artifacts including holes and visible disk-like structures. Similarly, we attempted to initialize the point cloud as 3D Gaussians (3DGS) to leverage recent differentiable Gaussian rasterization techniques~\cite{kerbl20233d}. However, the 3DGS renderer introduces an excessive number of parameters to optimize, which proved inefficient during the limited sampling steps of our diffusion process.  

\paragraph{Left-right consistency.} While left-right consistency checks are commonly employed in stereo methods~\cite{godard2017unsupervised}, we deliberately omit this approach in our framework. Unlike learning-based methods that can infer depth of both left and right from a single image, our optimization-based technique would require running the depth prediction process twice—once for each view—effectively doubling the computational cost. Therefore, in this work, we demonstrate our method's efficacy by optimizing the photometric loss using only a single reference view, achieving a favorable balance between accuracy and computational efficiency.

\subsection{Regularization}
\label{subsec:supp_reg}

\begin{figure*}[!htb]
\centering
\begin{minipage}{.45\textwidth}
  \centering
  \includegraphics[width=\linewidth]{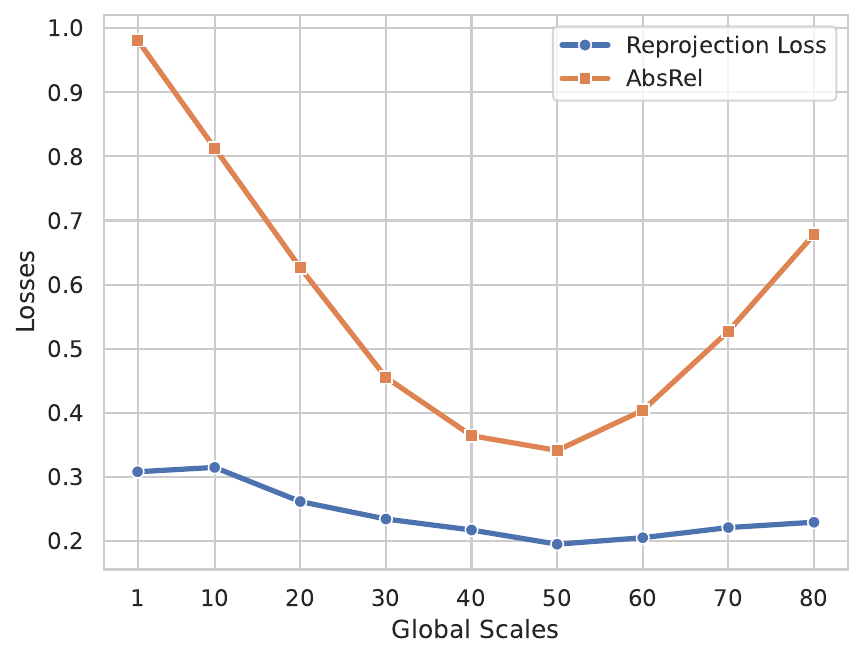}
  \captionof{figure}{Global scale up to $80$}
  \label{fig:lossAbs1}
\end{minipage}%
\begin{minipage}{.45\textwidth}
  \centering
  \includegraphics[width=\linewidth]{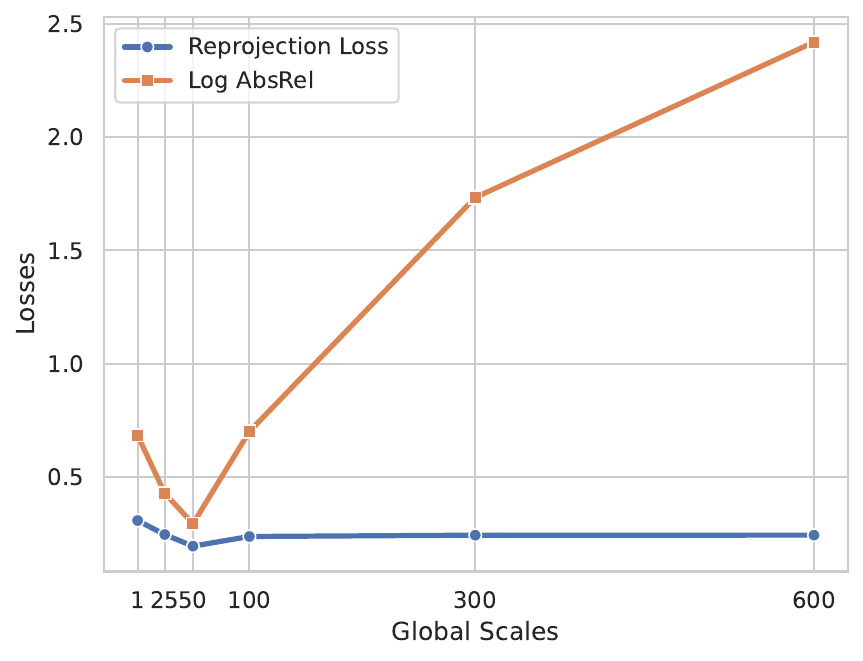}
  \captionof{figure}{Global scale up to $600$}
  \label{fig:lossAbs2}
\end{minipage}
\captionof{figure}{We gradually increase the global scale $g_s$ and observe a strong correlation between the reprojection loss and the AbsRel metric. For this example, the AbsRel reaches its minimum at a global scale of $50$. However, beyond $50$, the AbsRel significantly increases, while the reprojection loss shows little change, deviating from the pattern.}
\end{figure*}

\begin{figure*}[t]
  \begin{center}

  \includegraphics[width=\textwidth]{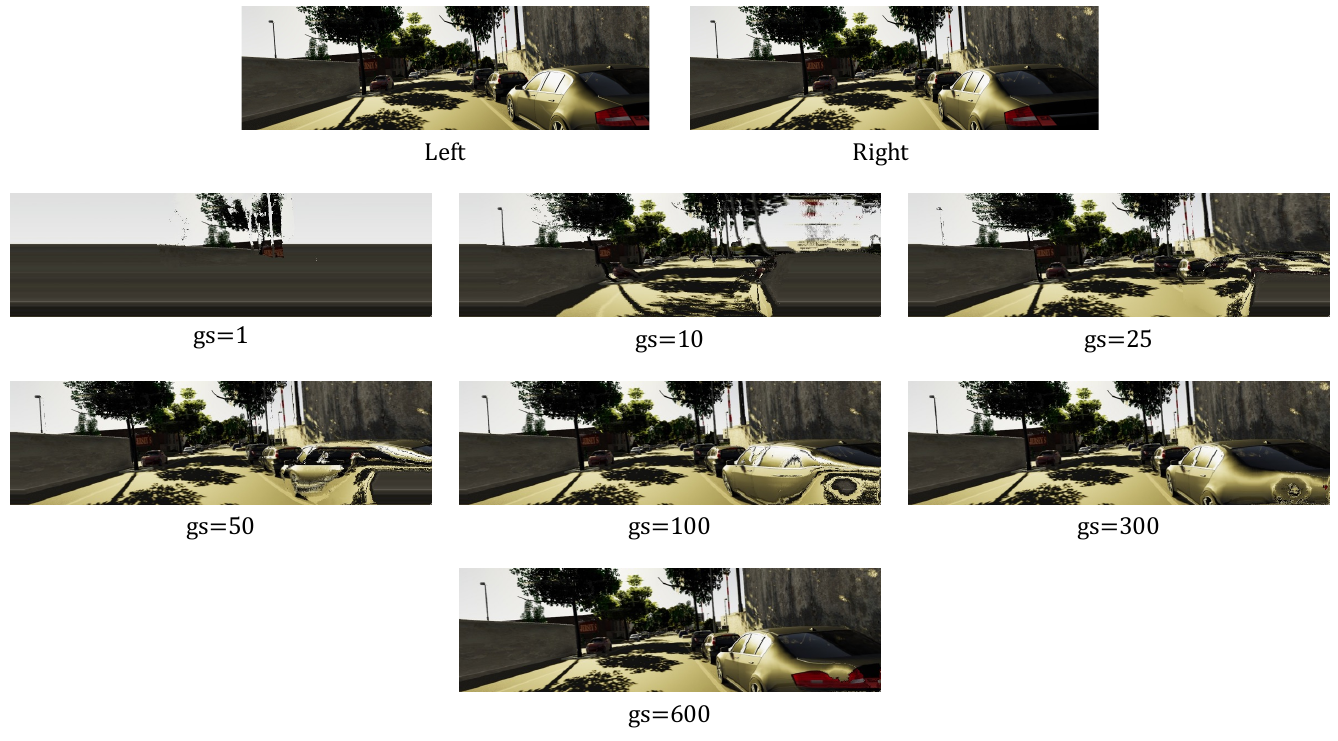}
  \end{center}
  \caption{We gradually increase the global scale $g_s$ and re-render the left image using the right image and $\tilde{\bm{x}}_0^{scale}$ (see Section~\ref{subsec:supp_reg}). While greater depths result in higher-quality re-rendered images, this does not necessarily correspond to more accurate predicted depths.}
  \label{fig:warp}

\end{figure*}

As described in Section~\ref{sec:method}, we stabilize the optimization process by introducing a global scale $ g_s $ and applying $ L_2 $ regularization to the scale $\hat{s}$ and shift $\hat{t}$ parameters. To illustrate the design of these parameters, we present a toy example. Given a predicted relative depth map $\tilde{\bm{x}}_0^{rel}$ (normalized to the range $[0,1]$), we incrementally increase the global scale $ g_s $ to compute the scaled depth $\tilde{\bm{x}}_0^{scale}$ such that $\tilde{\bm{x}}_0^{scale} = g_s \cdot \tilde{\bm{x}}_0^{rel}$. For each scale, we evaluate the Absolute Relative (AbsRel) metric (where lower values are better) using the scaled depth $\tilde{\bm{x}}_0^{scale}$ and the ground truth depth. Additionally, using the left image, right image, and the predicted scaled depth $\tilde{\bm{x}}_0^{scale}$, we compute the reprojection loss $\mathcal{L}_{geo}$, as defined in Equation~\ref{eq:geo-loss}, for each scale.

As shown in Figure~\ref{fig:warp}, increasing the depth scale improves the resemblance of the re-rendered image to the left image. This occurs because closer depths result in larger disparities between the source and target viewpoints, requiring more significant transformations to align the images. Such large transformations often cause distortions, stretching, or undersampling in areas lacking sufficient source information, degrading the quality of the re-rendered image. In contrast, at greater depths, disparities between the viewpoints are smaller, leading to less dramatic transformations. These smaller adjustments maintain spatial coherence more effectively and reduce interpolation artifacts, producing sharper and more accurate re-rendered images.

Figures~\ref{fig:lossAbs1} and \ref{fig:lossAbs2} demonstrate this pattern as the global scale $ g_s $ increases from $ 1 $ to $ 50 $. However, further increasing $ g_s $, while improving the re-rendered image quality and enhancing $\mathcal{L}_{geo}$, leads to worse AbsRel metrics. This indicates that the depth scale $\tilde{\bm{x}}_0^{scale}$ deviates from the ground truth depth. This behavior underscores the strong geometric guidance provided by $\mathcal{L}_{geo}$ for the diffusion model during sampling.

To balance these considerations, we pre-select $ g_s $ based on the geometric loss. Specifically, we search for $ g_s $ within a predefined depth range and define the optimal scale as $ g^\ast_s := \arg\min_{g_s} \mathcal{L}_{geo} $. Our approach can be viewed as a variant of the traditional Plane Sweep Volume technique~\cite{collins1996space}, commonly used in stereo vision. Unlike conventional methods, our approach leverages the predicted relative depth to identify the correct depth scale, which is then applied uniformly to all pixels.

Finally, we apply $ L_2 $ regularization to the scale and shift parameters of $\tilde{\bm{x}}_0^{rel}$ to counteract the tendency of the optimization process to inflate these parameters, which can lead to incorrect metric depth predictions.

\section{Algorithm}
\label{sec:supp_algorithm}
We provide detail pseudo algorithm for our method at Algorithm~\ref{algo}. To avoid confusion, note that while learnable scale and shift are denoted as ($\hat{s}, \hat{t}$), score function and time step of diffusion are denoted as ($\bm{s}, t$), respectively.

\begin{algorithm*}[!t]
\caption{Geometric-Guided Diffusion for Metric Depth Estimation}
\begin{algorithmic}
\label{algo}
\REQUIRE Stereo images $\bm{y}_1, \bm{y}_2$, camera intrinsics and extrinsics, pretrained diffusion model $\bm{s}_\theta(\bm{z}_t, t, \bm{y}_1)$
\STATE Initialize learnable scale $\hat{s}$ and shift $\hat{t}$
\STATE Initialize random noise $\bm{z}_T \sim \mathcal{N}(0, I)$.
\FOR{$t = T - 1$ to $0$}
    \STATE $\hat{\bm{s}}_{t+1} = \bm{s}_\theta(\bm{z}_t, t, \bm{y}_1)$ \hfill $\triangleright$ Compute the score
    \STATE $\hat{\bm{z}}_0=\frac{1}{\sqrt{\bar{\alpha}_t}}\left(\bm{z}_t+\sqrt{1-\bar{\alpha}_t} \bm{s}_\theta\left(\bm{z}_t, t, \bm{y}_1 \right)\right)$ 
    \hfill $\triangleright$ Compute relative depth using Tweedie’s formula
    \STATE $\tilde{\bm{x}}_0 = \text{softplus}(\hat{s}) \cdot \bm{D}(\bm{z}_0) + \text{softplus}(\hat{t})$
    \hfill $\triangleright$ Convert relative depth to metric scale

    \STATE Compute $\mathcal{L}_{geo}$ following Eq. $\ref{eq:geo-loss}$
    \STATE $\hat{s} \leftarrow \hat{s} - \lambda_{\hat{s}} \nabla_{\hat{s}} \mathcal{L}_{geo}$ \hfill $\triangleright$ Gradient update for $\hat{s}$
    \STATE $\hat{t} \leftarrow \hat{t} - \lambda_{\hat{t}} \nabla_{\hat{t}} \mathcal{L}_{geo}
    $ \hfill $\triangleright$ Gradient update for $\hat{t}$
    \STATE $ \bm{z}_{t-1} = \sqrt{\bar{\alpha}_{t-1}} \hat{\bm{z}}_0 + \sqrt{1 - \bar{\alpha}_{t-1}} \bm{s}_\theta(\bm{z}_t, t, \bm{y}_1) - \lambda \nabla_{\bm{z}_t} \mathcal{L}_{geo}$ 
    \hfill $\triangleright$ Perform DDIM step with geometric guidance
\ENDFOR
\STATE \textbf{Output:} Estimated metric depth map $\tilde{\bm{x}}_0$
\end{algorithmic}
\end{algorithm*}

\section{Implementation details}
\label{sec:supp_implementation_detail}
\begin{figure}
    \centering
    \includegraphics[width=\linewidth]{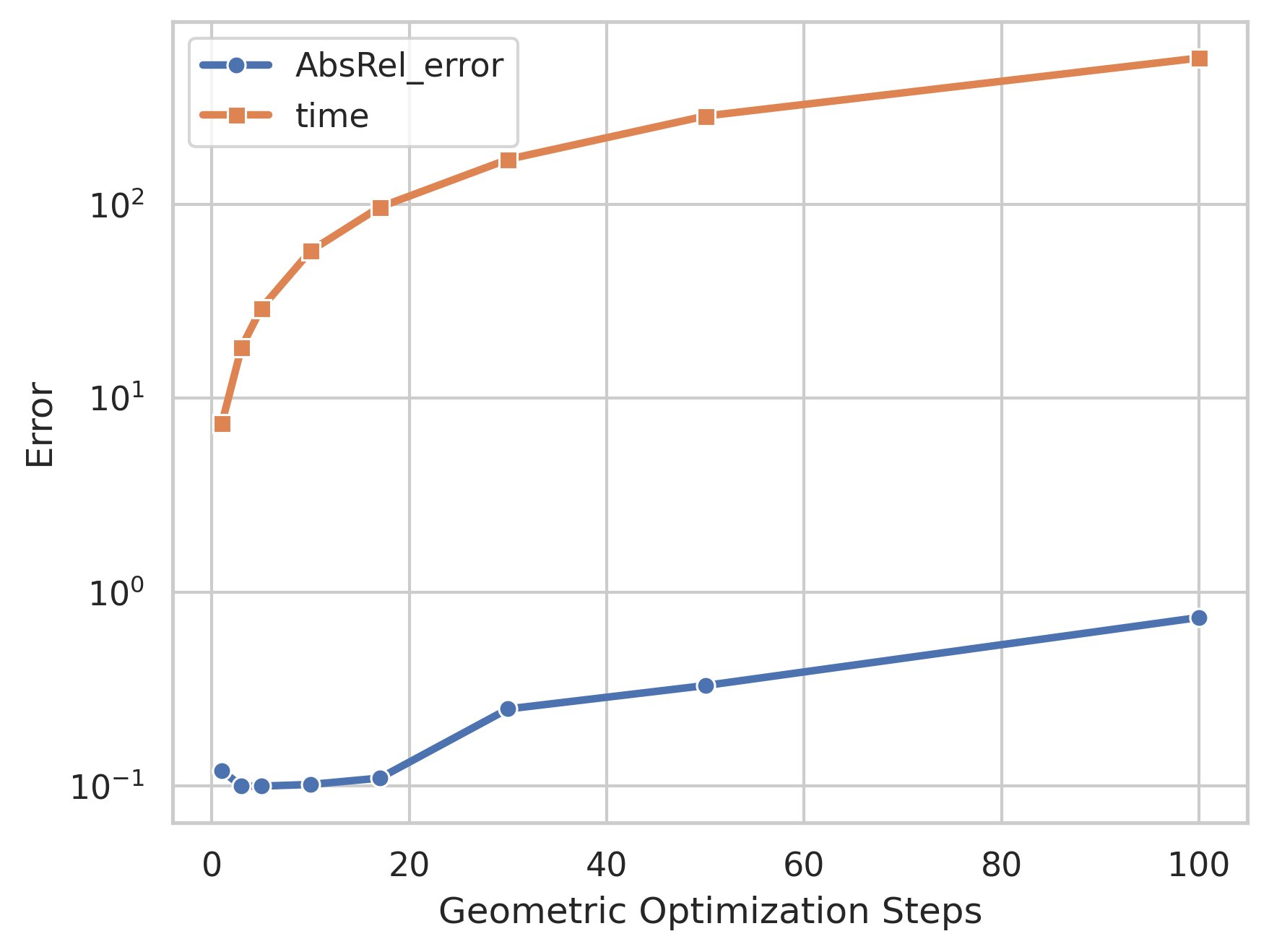}
    \caption{We increase the number of times to update inside one sampling step. We observe that it not only does not improve result, but also very time consuming to run.}
    \label{fig:error_time}
\end{figure}

\paragraph{Geometric optimization steps.} Our method employs a test-time optimization approach. While multiple gradient updates could theoretically be performed during sampling to minimize the geometric loss, our experiments in Figure~\ref{fig:error_time} demonstrate that only a limited number of gradient steps are beneficial. Consequently, we implement a single gradient update per sampling step in this work. Empirical observations indicate that increasing the number of gradient updates not only fails to improve performance but also significantly increases computational time.
\paragraph{Inference time.} Inference time for a single sample using our approach is approximately 7 seconds on an RTX A6000 GPU with images of 768 pixels in dimension. This measurement excludes data preparation time, which varies across datasets.  
\paragraph{Run time comparison.}  Our method requires test-time optimization but maintains computational efficiency comparable to the baseline Marigold~\cite{ke2024repurposing}, adding only seconds per image to processing time. This efficiency stems from our implementation of single-step gradient updates with minimal learnable parameters. Additionally, our depth warping-based rendering technique is both fast and fully differentiable. Consequently, despite achieving superior results, our approach does not significantly increase computational overhead compared to Marigold.

\section{Dataset details}
\label{sec:supp_dataset_detail}
\paragraph{Training data.} Our proposed method is a test time optimization-based, so we do not require any training sample. For details about training dataset of our baseline method, we refer reader to Marigold~\cite{ke2024repurposing}.
\paragraph{Evaluation data.} We evaluate our proposed approach on four distinct datasets. The KITTI-2015 dataset comprises 200 stereo pairs depicting outdoor scenes. The Middlebury dataset contains 15 stereo pairs predominantly featuring indoor environments. The Booster dataset includes 228 stereo pairs with challenging non-Lambertian surfaces. For the Tanks and Temples dataset, we randomly sampled 116 image pairs from a multi-view dataset spanning four scenes. 

\section{Limitation}
\label{sec:supp_limitation}
Since our method is based on diffusion sampling process, it is not suitable for real time application. Additionally, since we employ depth warping as a rendering technique and utilize photometric loss as an optimization objective, our approach exhibits sensitivity to significant illumination variations between stereo images. Potential solutions include applying color correction prior to image rendering or implementing left-right consistency as described in Section~\ref{subsec:supp_diff_warping}. We defer these improvements to future work.

\section{Additional qualitative results}
\label{sec:supp_qualitative}
\begin{figure*}[t]
  \begin{center}

  \includegraphics[width=\textwidth]{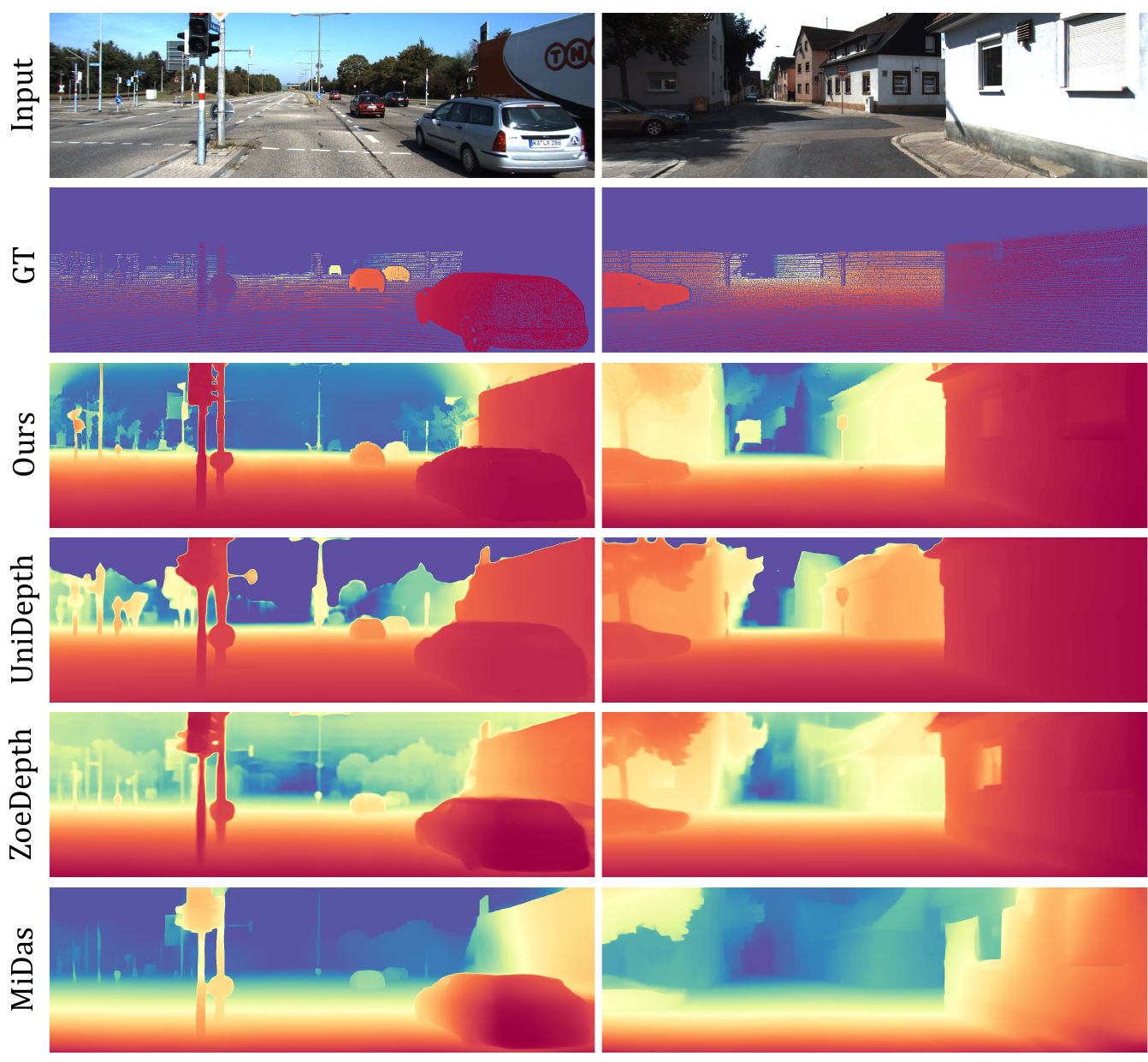}
  \end{center}
  \caption{\textbf{Additional qualitative results.} Our method demonstrates superior depth quality in metric depth estimation, particularly in high-depth-range scenarios such as those found in the KITTI-2015 dataset~\cite{menze2015joint}, outperforming competing approaches.}
  \label{fig:supp_kitti}

\end{figure*}

\begin{figure*}[t]
  \begin{center}

  \includegraphics[width=\textwidth]{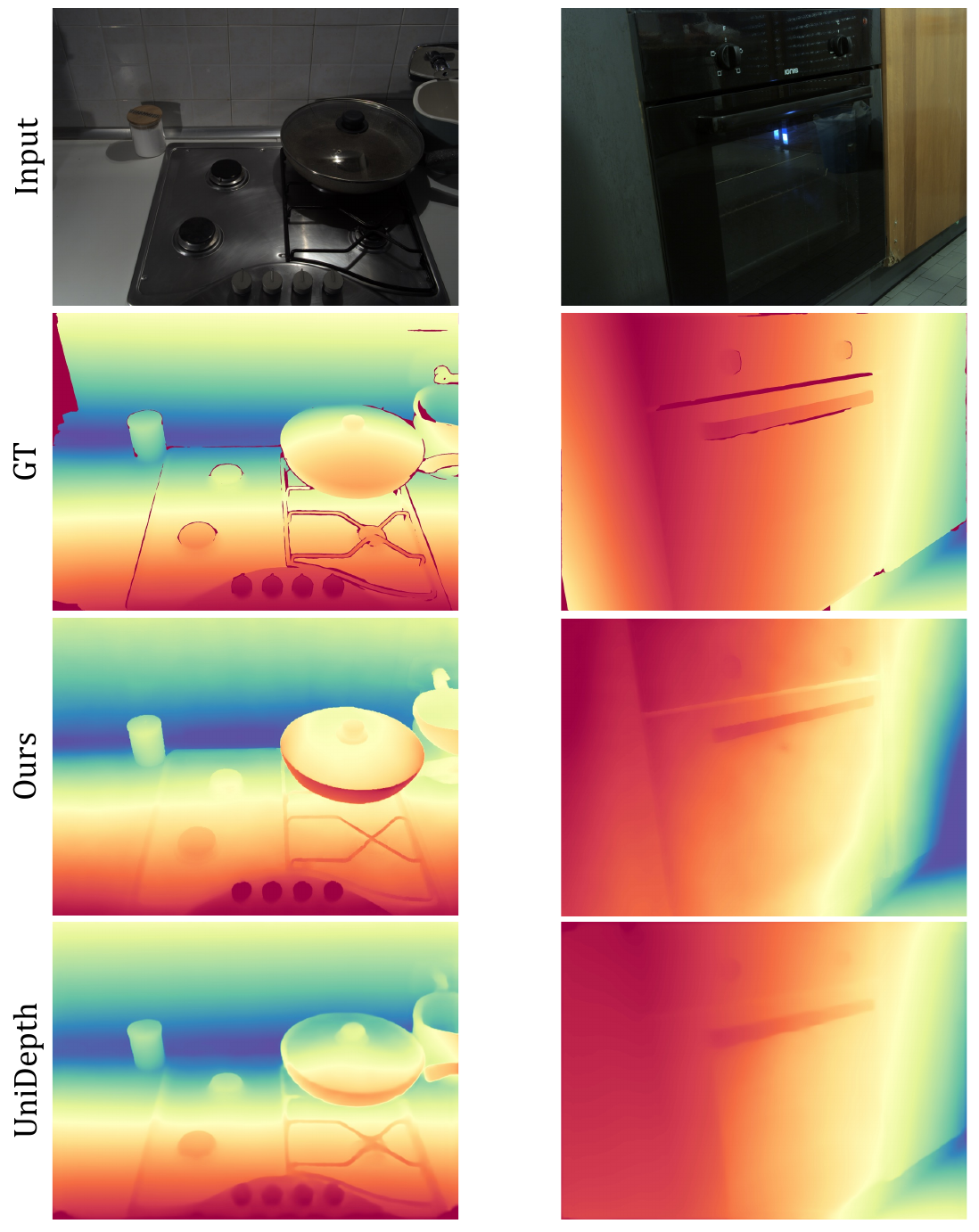}
  \end{center}
  \caption{Additional qualitative results on Booster dataset~\cite{zamaramirez2024booster}}
  \label{fig:supp_b1}

\end{figure*}

\begin{figure*}[t]
  \begin{center}

  \includegraphics[width=\textwidth]{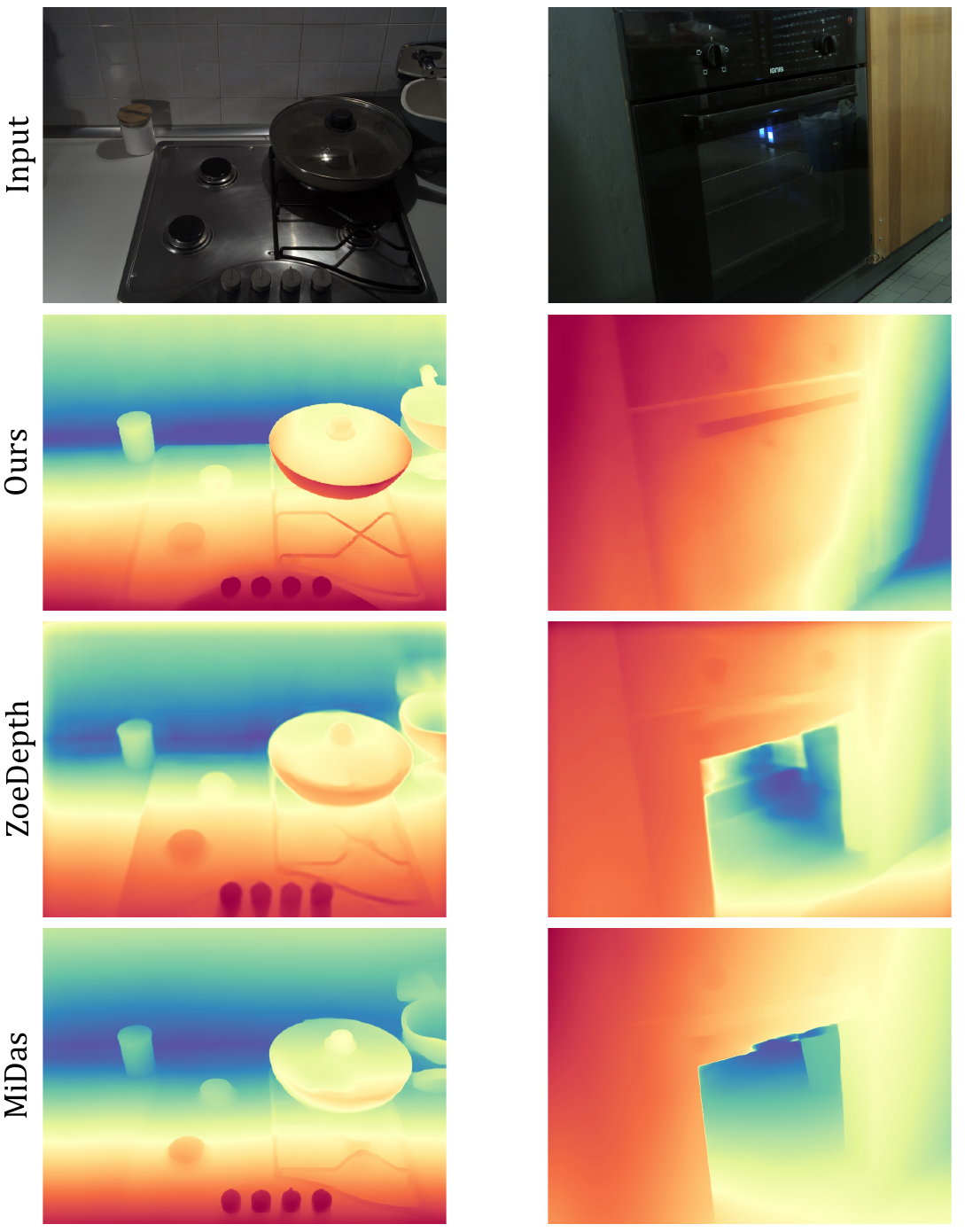}
  \end{center}
  \caption{Additional qualitative results on Booster dataset~\cite{zamaramirez2024booster}}
  \label{fig:supp_b2}

\end{figure*}

Additional qualitative results are presented in Figure~\ref{fig:supp_kitti}, Figure~\ref{fig:supp_b1}, Figure~\ref{fig:supp_b2}.

\newpage

\newpage

\clearpage
{
    \small
    \bibliographystyle{ieeenat_fullname}
    \bibliography{main}
}

\end{document}